# GIS Copilot: Towards an Autonomous GIS Agent for Spatial Analysis

Temitope Akinboyewa, Zhenlong Li*, Huan Ning, M. Naser Lessani

Geoinformation and Big Data Research Lab, Department of Geography, The Pennsylvania State University, University Park, USA

*zhenlong@psu.edu

**Abstract**

Recent advancements in generative artificial intelligence (AI) and particularly Large Language Models (LLMs), offer promising capabilities for spatial analysis. Despite their potential, the integration of generative AI with established GIS platforms remains underexplored. In this study, we propose a framework for integrating LLMs directly into existing GIS platforms, using the well-developed QGIS software as an example. Our approach leverages the reasoning and programming capabilities of LLMs to autonomously generate spatial analysis workflows and code through an informed agent that has comprehensive documentation of key GIS tools and parameters. The framework also incorporates external tools such as GeoPandas to enhance the system's geoprocessing capabilities. The implementation of this framework resulted in the development of a "GIS Copilot" that allows GIS users to interact with QGIS using natural language for spatial analysis. The GIS Copilot was evaluated with over 100 spatial analysis tasks with three complexity levels: basic tasks that require one GIS tool and typically involve one data layer to perform simple operations; intermediate tasks involving multi-step processes with multiple tools, guided by user instructions; and advanced tasks which involve multi-step processes that require multiple tools but not guided by explicit instructions, necessitating the agent to independently decide on and executes the necessary steps. The evaluation reveals that the GIS Copilot demonstrates strong potential in automating foundational GIS operations, with a high success rate in tool selection and code generation for basic and intermediate tasks, while challenges remain in achieving full autonomy for more complex tasks. This study contributes to the emerging vision of autonomous GIS, providing a pathway for non-experts to engage with geospatial analysis with minimal prior expertise. While full autonomy is yet to be achieved, the GIS Copilot demonstrates significant potential for simplifying GIS workflows and enhancing decision-making processes.

# 1 Introduction

Recent advancements of generative artificial intelligence (AI), particularly the Large Language Models (LLMs), have gained significant attention in geographic information science (GIScience) through their abilities in reasoning, natural language understanding, and code generation. LLMs offer a wide range of potential applications within geographic information systems (GIS), including spatial data analysis, geoprocessing automation, and geospatial knowledge extraction. A pioneering study by Li and Ning (2023) introduced autonomous GIS as a next-generation AI-powered GIS, demonstrating the potential of generative AI in generating geoprocessing workflows and conducting spatial analysis autonomously. A number of studies have further examined the integration of generative AI (LLMs) within GIS. For instance, researchers have explored its applications in cartography (Tao and Xu, 2023; Zhang et al., 2024a), disaster management (Akinboyewa et al., 2024; Hao et al., 2024), GIS-based question answering (Mooney et al., 2023; Zhang et al., 2024b), location descriptions extraction (Hu et al., 2023), and geospatial data retrieval (Ning et al., 2024). These efforts collectively demonstrate that generative AI holds promise in enhancing and potentially transforming traditional GIS workflows.

While the integration of generative AI in GIS has shown significant promise, one key limitation is that these existing approaches have not fully embedded LLM capabilities within established GIS platforms like QGIS (2024), ArcGIS Pro (ESRI, 2024), and GRASS GIS (2024). Li and Ning (2023) suggest that implementing autonomous GIS based on existing GIS platforms as a "co-pilot" is likely the most practical and efficient approach, as established GIS platforms already provide extensive tools for spatial analysis, data management, and visualization. However, the integration of LLMs into GIS platforms like QGIS presents several challenges. First, GIS platforms are inherently complex, supporting diverse tools, libraries, and workflows for data visualization, spatial analysis, geoprocessing, and cartographic rendering. These tools often rely on diverse frameworks, data standards, and programming languages, complicating the creation of a single, cohesive workflow that integrates AI smoothly. Second, effective communication between the AI and existing GIS functions requires robust and error-free interaction. This involves translating natural language queries into GIS commands and ensuring that those commands align with GIS native tools and associated tool parameters. Such a process demands extensive fine-tuning and specialized approaches to understand the full breadth of GIS tools and operations. Additionally, GIS workflows are often highly customized and problem specific. Developing an AI integration that can seamlessly adapt to these custom workflows, while being flexible enough to handle various user-defined parameters, is challenging.

To address these challenges, we propose a framework to integrate generative AI into a well-established GIS platform that leverages the advanced capabilities of LLMs to interact with existing geoprocessing tools. To implement this framework, we developed a user-friendly plugin that serves as a "Copilot" in the QGIS software, allowing users to perform spatial analysis through natural language queries. This GIS Copilot provides a graphic user interface where users can input a task (spatial problem), which the agent analyzes by breaking it into steps and selecting suitable GIS tools. It then generates and executes Python code for the task to produce the final result. By leveraging the existing capabilities of GIS platforms and enhancing them with AI-driven reasoning, this framework aims to overcome the limitations of previous efforts and contribute to the advancement of autonomous GIS research.

The key contributions of this study can be summarized as follows: 1) This study introduces an extensible and generalizable framework for integrating LLMs into established GIS platforms, setting a precedent for future developments in GIS Copilot and autonomous GIS across multiple platforms (e.g., ArcGIS, GRASS GIS). 2) This study creates pathways for GIS accessibility and adoption. By enabling natural language interaction and expanding GIS capabilities to non-expert users, this research demonstrates how AI-driven systems can potentially transform geospatial analysis in non-traditional GIS domains like public health, disaster response, and urban planning. 3) To the best of our knowledge, this is the first GIS Copilot deeply integrated into a well-established GIS platform, supporting diverse geospatial tasks while enhancing transparency, real-time reasoning, and expandability for external tools. While full autonomy is not yet achieved, it brings us closer to realizing autonomous GIS systems for complex geospatial tasks with minimal human intervention.

The remainder of this paper is organized in this manner. Section 2 reviews related work on autonomous GIS and the integration of generative AI into geospatial tasks. Section 3 details the proposed framework and prototype workflow. Section 4 presents the results from case studies and evaluation of the performance of the Copilot at different complexity levels of cases. Section 5 discusses the key findings and Section 6 discusses the limitations and potential future research directions. Finally, Section 7 concludes the paper.

## 2 Related Work

### 2.1 Autonomous GIS and the integration of generative AI into geospatial tasks

The recent advancement of generative AI, particularly LLMs such as GPT-4 (OpenAI, 2023), Llama (Meta, 2024), and Gemini (Gemini Team and Google, 2023), has opened the avenue for researchers to explore autonomous GIS (Li and Ning, 2023; Wang et al., 2024). Researchers have leveraged LLMs to conceptualize and develop various autonomous GIS agents from geospatial data retrieval (Ning et al., 2024) to geospatial information extraction (Crooks and Chen, 2024) to spatial analysis and mapping (Mansourian and Oucheikh, 2024; Zhang et al., 2024b). Among the pioneering studies in autonomous GIS is the work by Li and Ning (2023), who developed a prototype named LLM-Geo to demonstrate the potential of AI-powered GIS to perform various geospatial operations such as numerical aggregation, chart generation, and map presentations in an autonomous manner using GPT-4 API. In another study, Zhang et al. (2024b) developed GeoGPT, which integrates the capabilities of LLMs with GIS tools to enhance the efficiency of geospatial operations, including data collection, processing, and analysis. By utilizing the LangChain framework, the LLM interprets user requests, plans the necessary steps, and sequentially executes relevant GIS tools from a predefined tool pool, encompassing tools for data collection, spatial analysis, and cartographic visualization. Additionally, Lin et al. (2024) recently proposed a framework designed to automate shapefile processing tasks using a multi-agent approach.

Furthermore, several other studies have integrated LLMs into specific spatial tasks. For instance, Tao and Xu (2023) explored the integration of ChatGPT into cartography to automate and enhance map-making tasks. Their study conducted experiments on generating thematic and mental maps by using natural language prompts in ChatGPT, which produced Python codes for geospatial visualization. In a relevant study, Zhang et al. (2024a) introduce MapGPT, an LLM-based mapping framework that integrates GPT-4 with multiple mapping tools that allow users to interact through natural language descriptions, simplifying the mapping process. Mai et al.

(2024) examine the potential of foundation models, including LLMs, for geospatial artificial intelligence (GeoAI). They evaluate the performance of existing pre-trained LLMs on a range of geospatial tasks, such as toponym recognition and location description, as well as multimodal tasks like urban use classification and remote sensing image analysis. Xu et al. (2024) developed a comprehensive benchmark to evaluate large language models (LLMs) on spatial tasks. While these studies revealed the potential of generative AI in geospatial operations, one of the key limitations is that these approaches have not fully integrated generative AI capabilities within established GIS platforms.

## 2.2 Integrating generative AI into existing GIS platforms

Recent efforts have been made to integrate generative AI with established GIS platforms. Ning et al. (2024) introduced an autonomous GIS agent framework and a QGIS plugin to facilitate the automated downloading of geospatial data. While this study represents significant progress, it focuses primarily on data retrieval rather than the broader spectrum of spatial analysis tasks. QChatGPT (2023) is another LLM-based QGIS plugin, but its functionality is currently limited to offering guidance on performing tasks rather than automating the tasks themselves. IntelliGeo (2024) is a QGIS plugin that leverages LLMs to automate geospatial workflows. It is designed to assist GIS professionals by automatically performing GIS operations through two primary methods: generating PyQGIS scripts or creating graphical models. Although promising, it is limited to generating general code, requiring users to manually configure parameters in the code, reducing the automation level.

Recently, Mansourian and Oucheikh (2024) introduced ChatGeoAI which utilizes a fine-tuned Llama2 model to generate code that can be used to perform spatial tasks. While this approach shares the goal of creating a platform that integrates LLM with GIS for geospatial analysis tasks, our study differs in several ways. First, our study proposes a framework that deeply integrates LLMs within a well-established platform like QGIS, leveraging the full range of GIS tools. This allows the LLM to be more informed and produce more accurate and context-specific code. Second, our framework is implemented through the development of a GIS Copilot that is fully integrated into the GIS software itself to support a wide range of operations across various data types, including vector, raster, and attribute data in tabular files, which offers greater flexibility in geospatial analysis. Third, the design of the GIS Copilot enhances transparency by providing explanations of the tool selection process and a real-time display of the reasoning process and generated code, offering users insights into the execution steps. Additionally, our framework is not limited to QGIS tools alone; it is designed to be expandable, allowing the integration of external libraries and tools. For example, the widely used Python package, *seaborn*, can be used for advanced charting, making certain tasks more efficient. Lastly, we incorporate a self-correcting mechanism aimed at minimizing errors to enhance the system's autonomy and adaptability during task execution.

ESRI, a leading global provider of GIS software, has also taken advances toward integrating generative AI into ArcGIS, introducing AI-enhanced tools aimed at improving geoprocessing workflows (Crawford et al., 2024). This includes functionality that recommends suitable tools for specific operations, providing guidance but not automating tasks. Additionally, ESRI's AI Assistant allows users to initiate GIS tasks through natural language inputs (ESRI, 2024a). However, this development is still in early stage at the time of writing. ESRI also introduced Copilot for Microsoft 365 that allows users to search, access maps, apps, and geospatial data within ArcGIS for Teams (ESRI, 2024b). More recently, NASA and Microsoft have collaborated to develop Earth Copilot, an AI-driven tool designed to simplify and democratize access to NASA's extensive Earth science data through natural language queries, such as assessing the impact of specific events or environmental changes (Microsoft, 2024).

## 3 Methodology

### 3.1 Framework of the GIS Copilot

The GIS Copilot is designed to interact with existing GIS platform (e.g., QGIS) via graphic user interface, GIS toolbox, and documents for the toolbox, to assist GIS users in conducting geospatial data processing and analysis tasks. The framework consists of four core modules, including data understanding, GIS interface interaction, code review/debugging, and tool documentation. The GIS Copilot is informed by the data insights from the data understanding module and leverages documentation of the geoprocessing tools contained in the spatial analysis toolbox, to conduct spatial analyses. It automatically determines the most suitable geospatial processing tools for various tasks, manages data processing workflows, and executes spatial analysis operations. It generates the necessary code for the selected tools and uses the code review and debugging module to automatically correct errors that may arise during execution.

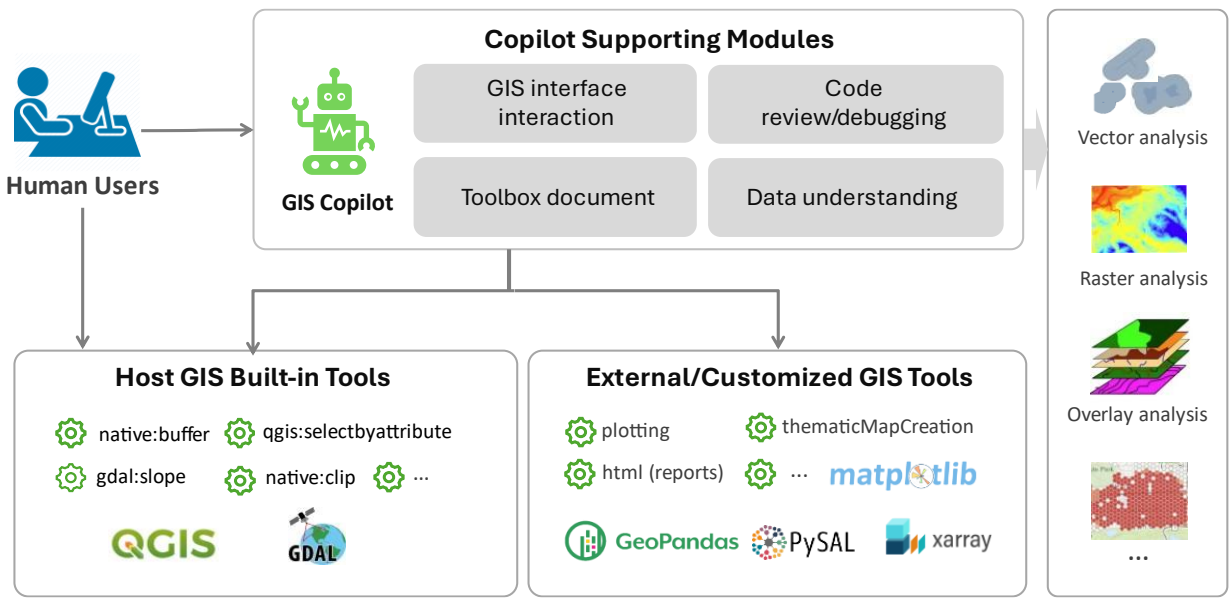

Figure 1: Framework of the GIS Copilot

#### 3.1.1 Spatial analysis toolbox

QGIS is a widely used open-source GIS software for geospatial data management, analysis, and visualization. It offers tools from various algorithm providers, including QGIS native algorithms, GDAL/OGR, GRASS GIS, SAGA GIS, and Orfeo Toolbox, supporting diverse geospatial workflows. Additionally, its Python API, PyQGIS, allows users to automate tasks, create custom plugins, and access full QGIS functionality, making it highly adaptable for advanced spatial analysis and tool development (QGIS, 2024b).

The spatial analysis toolbox in this framework comprises both built-in and customized geoprocessing tools. Built-in tools include QGIS algorithms, such as native geoprocessing algorithms (e.g., 'native:buffer' for buffering and 'qgis:selectbyattribute' for attribute selection) and external providers like GDAL (e.g., 'gdal:slope' for slope analysis). The customized geoprocessing tools, developed using external libraries like GeoPandas, PySAL, and Matplotlib. These customized tools enable statistical analysis (e.g., correlation analysis), customized charts (e.g., bar plot, scatter plot), and thematic map generation, offering flexibility in spatial analysis workflows.

### 3.1.2 Toolbox documentation

The GIS Copilot is equipped with documentation of QGIS tools, allowing it to generate PyQGIS code for specific geoprocessing tasks. Derived from the QGIS help documents, the toolbox documentation primarily consists of tool descriptions, parameters, and code examples, serving as a guide for the agent in generating code for selected tools. For instance, the QGIS native "Clip" tool requires parameters such as the Input Layer (the dataset to be clipped), Overlay (the boundary layer for clipping), and Output File (the path for saving the clipped data) (Appendix 1). By specifying these parameters and providing a code example, the system is able to generate more accurate and reliable code for the task.

In our current implementation, 390 native QGIS and GDAL tools have been systematically compiled with reference to the resources available on the official QGIS documentation website (QGIS, 2024b). In addition, this framework is not limited to QGIS tools and is designed to be extensible in a plug-and-play style to allow users to add new customized tools to meet specific tasks. For example, a tool used for performing kernel density estimation can be customized using *geopandas*, *scipy*, and *rasterio* Python libraries and registered to the GIS Copilot (Appendix 2).

### 3.1.3 Data understanding module

The data understanding module enables the agent to extract and interpret the structure and attributes of spatial data before being used for any analysis. It functions similarly as the initial step a human analyst would take when approaching a spatial analysis task -understanding the data to be analyzed. Specifically, it collects a wide range of critical information of the data for spatial analysis, such as file format, field names of the attribute table, coordinate reference system, data extent, and raster resolution. This information provides the GIS Copilot with a thorough overview and understanding of the data inputs, which guides it in selecting appropriate tools for the task and the right attributes for performing geospatial operations. As a result, users are not required to specify data details such as field names in their task requests, making the Copilot more user-friendly.

### 3.1.4 Code review/debugging module

The code review/debugging module is an essential component designed to enhance the reliability and accuracy of generated code. This module enables the GIS Copilot to actively review the generated code and monitor the execution of the generated code in real-time to capture any errors that may arise. Upon detecting an error, it attempts to correct the error automatically and re-generate the code for execution. This self-debugging functionality is iterative, allowing for continuous refinement and improvement of the code to increase the chances of successful execution.

### 3.1.5 GIS interface interaction

The GIS interface interaction module enables the GIS Copilot to engage directly with the host GIS' (QGIS) graphic user interface. It allows the Copilot to perform actions such as dynamic loading and data handling within the GIS software. For example, data layers that users load into GIS software can be automatically accessed by the Copilot. Additionally, results generated from the Copilot are directly reloaded into the host GIS, which allows immediate visualization and further processing or manipulation by the user.

## 3.2 Workflow of the GIS Copilot

The GIS Copilot is implemented as a plugin within QGIS. It interprets user requests in natural language and translates them into executable geoprocessing commands. As shown in Figure 2, the overall workflow begins by breaking down the user's task into detailed steps. Once the task is well understood, the system identifies the appropriate geoprocessing tool(s) from a wide range of options, including QGIS native tools, GDAL, or customized tools.The tools are selected based on the steps analyzed during the task breakdown to ensure each step of the process is addressed by the most suitable tool. Based on the selected tool (s), the system generates executable Python code that is specifically designed to solve the spatial task. Throughout these steps, the system is guided by a set of predefined instructions, automatically generated metadata and a comprehensive tool list to ensure reliable task analysis, tool selection, and code generation. Finally, the code is executed, and the results are either loaded as map layers into QGIS or presented as reports including tables and charts.

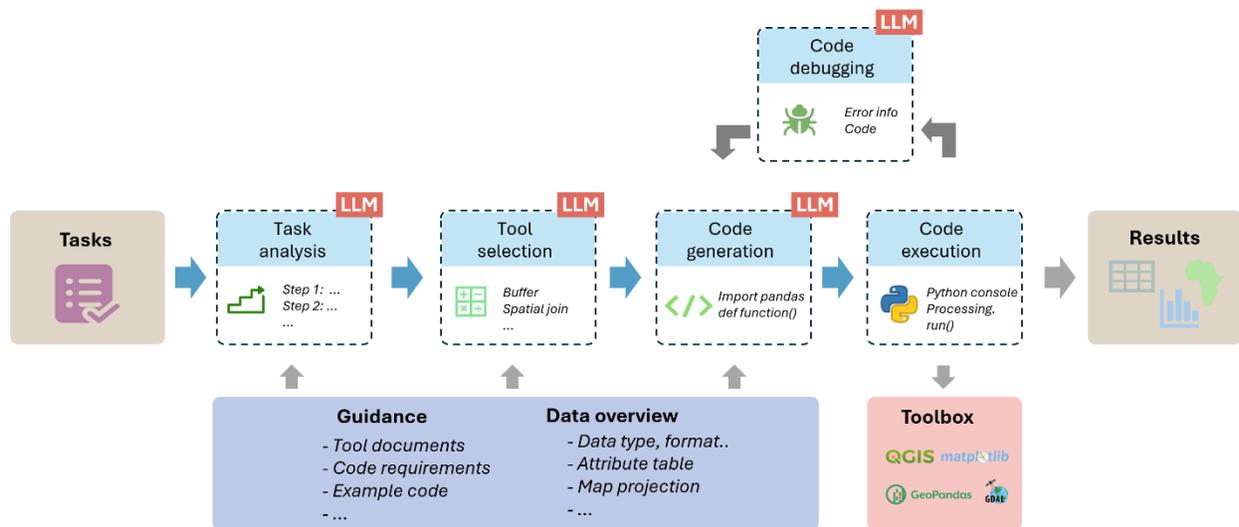

Figure 2: Workflow of the GIS Copilot

### 3.2.1 Task analysis and tool selection

When a user submits a spatial query in natural language, the first step is that the agent processes the request to understand the underlying spatial problem. After the initial interpretation, the agent performs a task breakdown to determine what is needed to achieve the intended outcome. For instance, a user may request, "Identify all residential areas within a 5-minute driving distance from hospitals." The agent first identifies the spatial entities involved - residential areas, hospitals, and the concept of a 5-minute drive. Based on these insights, it matches the request to a set of geoprocessing tools available within the spatial analysis toolbox. The decision-making process for selecting the appropriate tool involves considering several factors, such as the nature of the data (e.g., vector or raster), the geographic projection, and the data structure (e.g., polygons, lines). Additionally, any user preferences specified during the query, such as the choice of tools or specific processing techniques, are taken into account.

### 3.2.2 Code generation and parameter setting

Once the appropriate tools are selected, the next step in the workflow is the code generation phase. The agent generates Python code to execute the operation based on the selected tools. The generation process is guided by predefined documentation for each tool, which includes usage guides, parameter requirements, and sample code templates so that the generated code is well-formed and ready for execution. To personalize the generated code to fit the user's requirements, the agent adjusts the parameters to align with the task's specific details. For instance, suppose a user request is to "select counties with a population greater than 50,000". The agent customizes the code based on the user's specific requirement by setting the filter condition to "Population" > 50,000, where "population" is assumed to be the relevant attribute name.

### 3.2.3 Execution and visualization of results

After generating the code, the next step involves executing it to perform the spatial analysis. This stage is facilitated by an execution and self-debugging module, which oversees the entire process. The generated code is executed directly within the host GIS environment, allowing the spatial analysis to be performed without leaving the QGIS interface. Depending on the nature of the task, results can take various forms: the outputs of geoprocessing tasks, such as buffer zones, intersected layers, or reclassified raster layers, are loaded and displayed directly in QGIS as map layers, where they can be further analyzed, visualized, or exported. Analytical results that involve summaries, charts, attribute statistics, or calculated metrics are displayed within the plugin interface as tables, images, or textual reports. When appropriate or requested in the tasks, the plugin may also generate visualizations, such as bar charts, histograms, or other graphical representations to help users interpret the results. If an error is encountered during execution, the self-debugging module is activated. This module reviews the code, identifies the errors, makes necessary corrections, and re-executes the revised code.

### 3.3 User interface design

The graphic user interface of this GIS Copilot comprises six main tabs: Request Page, Generated Code, Geoprocessing Workflow, Reports, Settings and Help. The Request Page is the main interface where users can enter their requests. It includes key features like an AI-generated code displayed in real-time, an information panel showing the agent's status, a request box for user requests, and paths for datasets loaded into QGIS. To enable seamless integration with QGIS, users can select datasets for analysis directly from the QGIS layer panel by checking the layers they want to include and unchecking those to exclude. The Generated Code tab enables users to manually execute the code generated by the agent and make edits as needed. This tab also provides a platform to load saved code from the local machine or save generated code for future use. The geoprocessing workflow tab illustrates the steps involved in completing the user's request in a visual flow. Reports tab allows users to view outputs like charts, plots, and statistical summaries. The Settings tab lets users configure the plugin, such as setting API keys, selecting the AI model, and defining the default workspace where results are saved. Help tab offers the users the details on how to use the plugin.

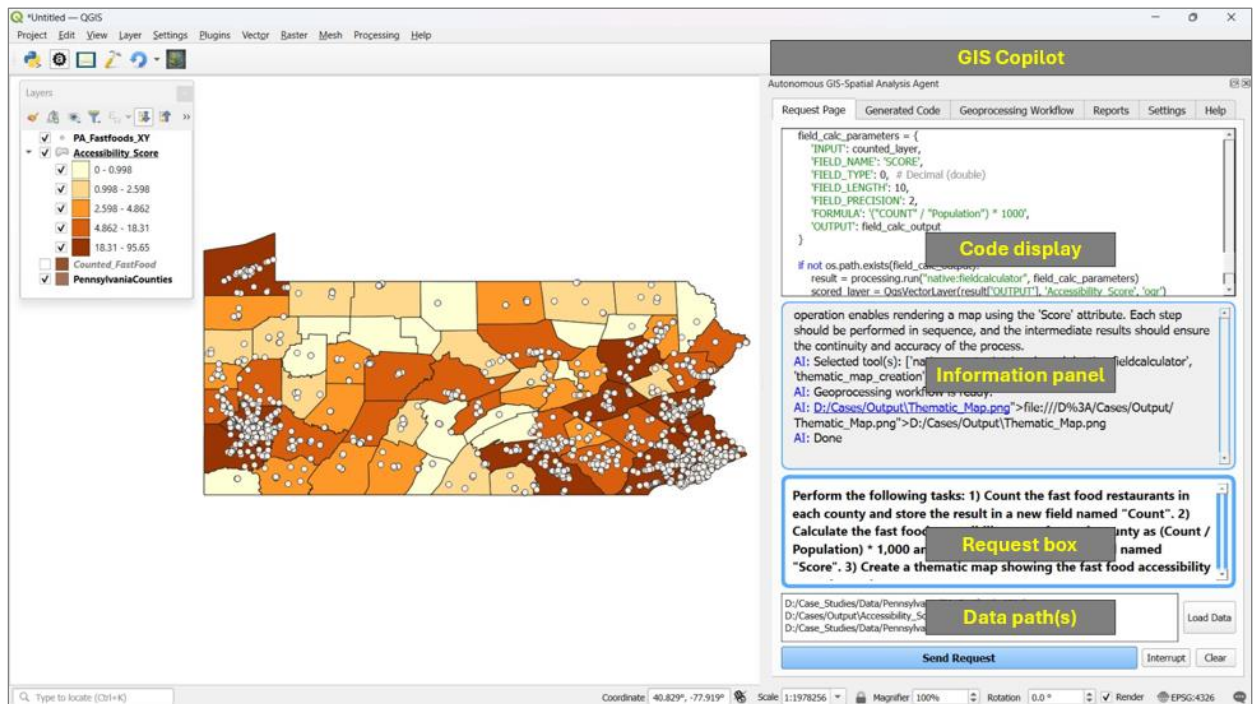

Figure 3: The graphic user interface of the GIS Copilot

## 4   Case Studies

To demonstrate the capabilities of the GIS Copilot, cases are categorized into three levels of complexity: basic, intermediate, and advanced.

### 4.1   Basic level tasks

At this level, the tasks are straightforward, involving a single tool along with one (or occasionally two) data layers. The agent is expected to perform a straightforward operation that usually requires a single step, such as calculating the area of polygons or selecting features based on attributes.

#### 4.1.1   Health facilities coverage zone in Washington D.C., USA

This case involves creating service zones around health facilities in Washington D.C, United States. The input data was a vector layer of health facilities (hospitals, clinics, and pharmacies) obtained from OpenStreetMap. In this task, the agent was prompted to create 2,000-foot zones around each health facility to identify areas of service coverage (Figure 4). The agent successfully selected the appropriate spatial analysis tool (the buffer tool) and generated executable code, which includes setting the required parameters (e.g., distance) based on the user query and executing the operation.

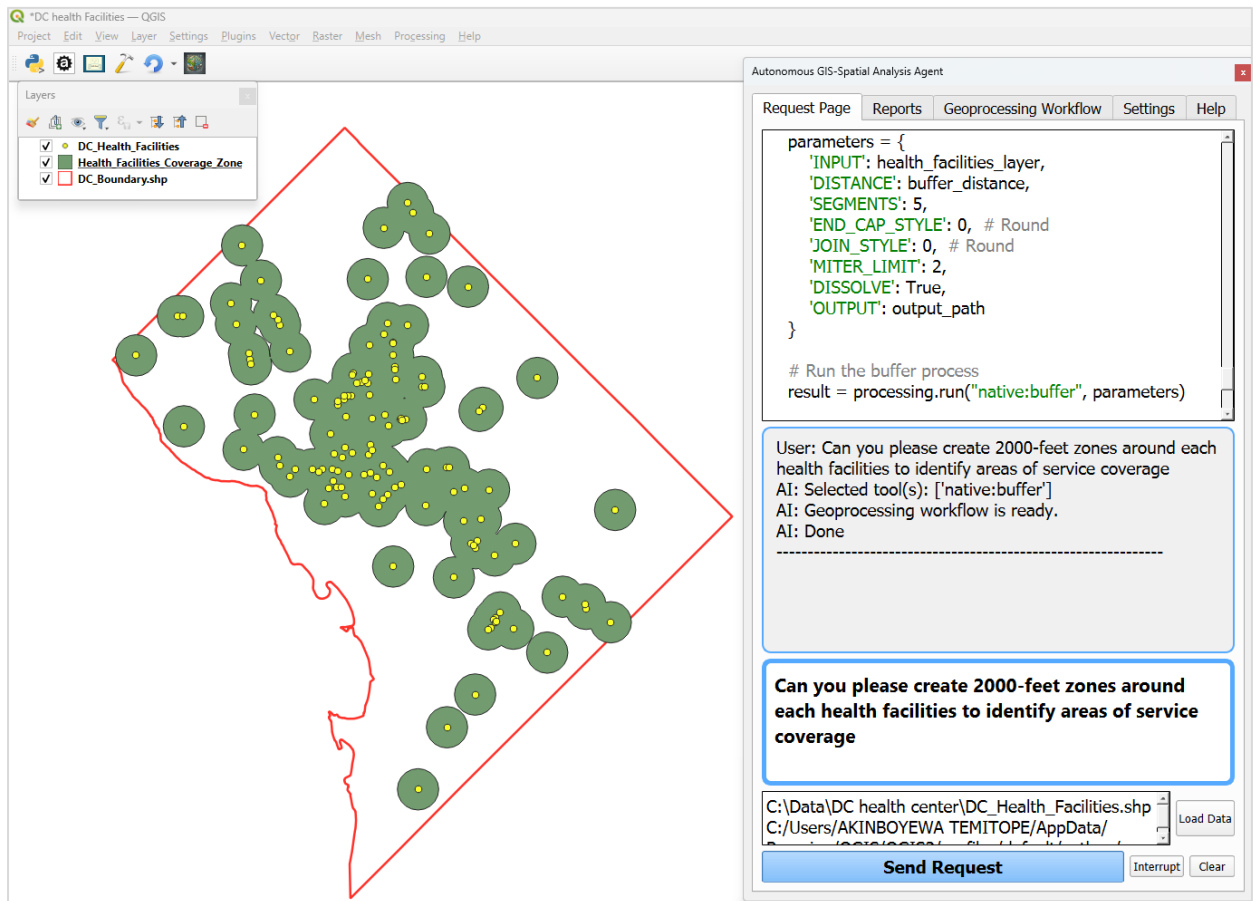

Figure 4: Health facilities coverage zone (Task: *Can you please create 2000-feet zones around each health facilities in Washington DC to identify areas of service coverage?*)

### 4.1.2 Generating contour lines from the Digital Elevation Model (DEM)

In this case, the agent is tasked with generating contour lines from a Digital Elevation Model (DEM) dataset, a raster data layer that represents terrain elevation. This is a foundational spatial analysis task that is frequently used in geographic studies, such as topographical mapping, environmental modeling, and landscape visualization. The agent selected the tool *gdal:contour* from the available geoprocessing tools within the system. This tool is specifically designed to convert elevation data (raster) into contour lines (vector), based on user-defined intervals. In our case study, the output in vector format is shown in Figure 5.

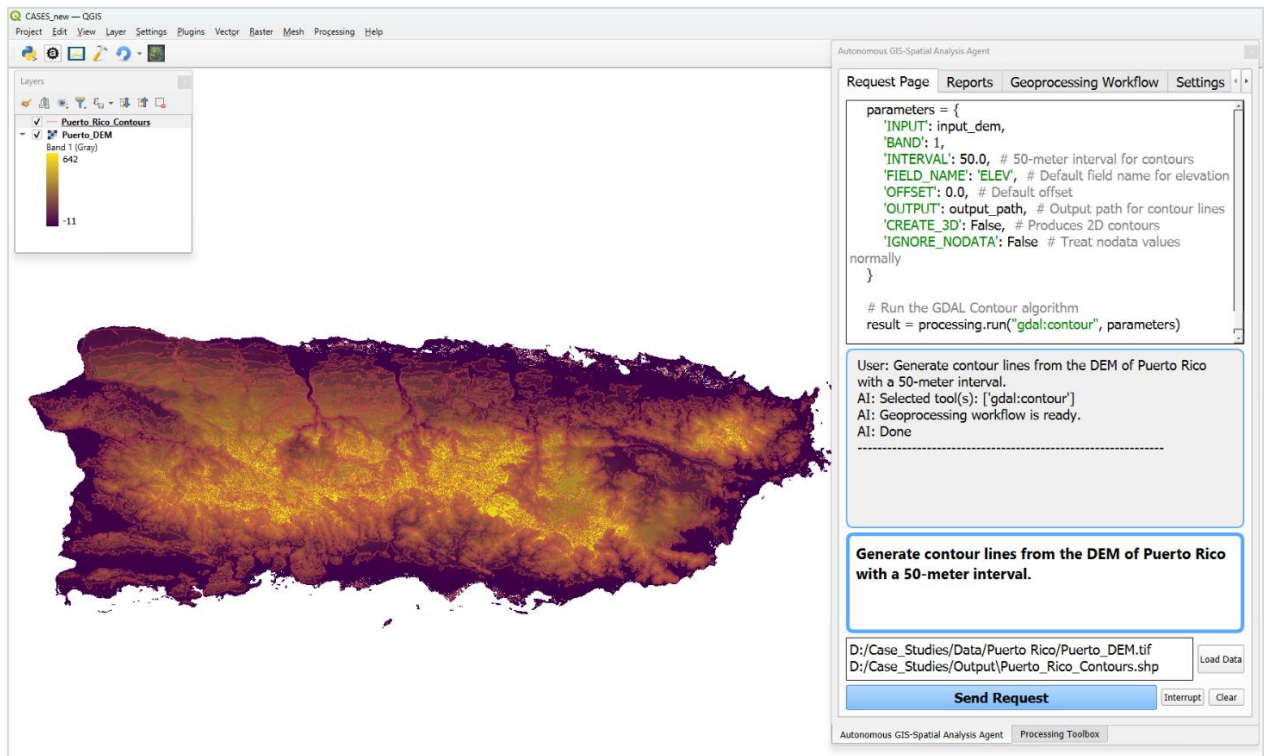

Figure 5: Generating contour lines from DEM (Task: *Generate contour lines from the DEM of Puerto Rico with a 50-meter interval.*)

### 4.1.3  Selection of high population counties in the US.

This case demonstrates attribute query by selecting the US counties with a population exceeding 50,000. The input dataset is a single vector layer that contains the population of all counties in the contiguous US states. In response to the task, the agent effectively identified the correct tool (extract by attribute) to filter the data. The agent then generated the necessary executable code, which included setting the population threshold and applying the filtering operation based on the specified condition (Figure 6).

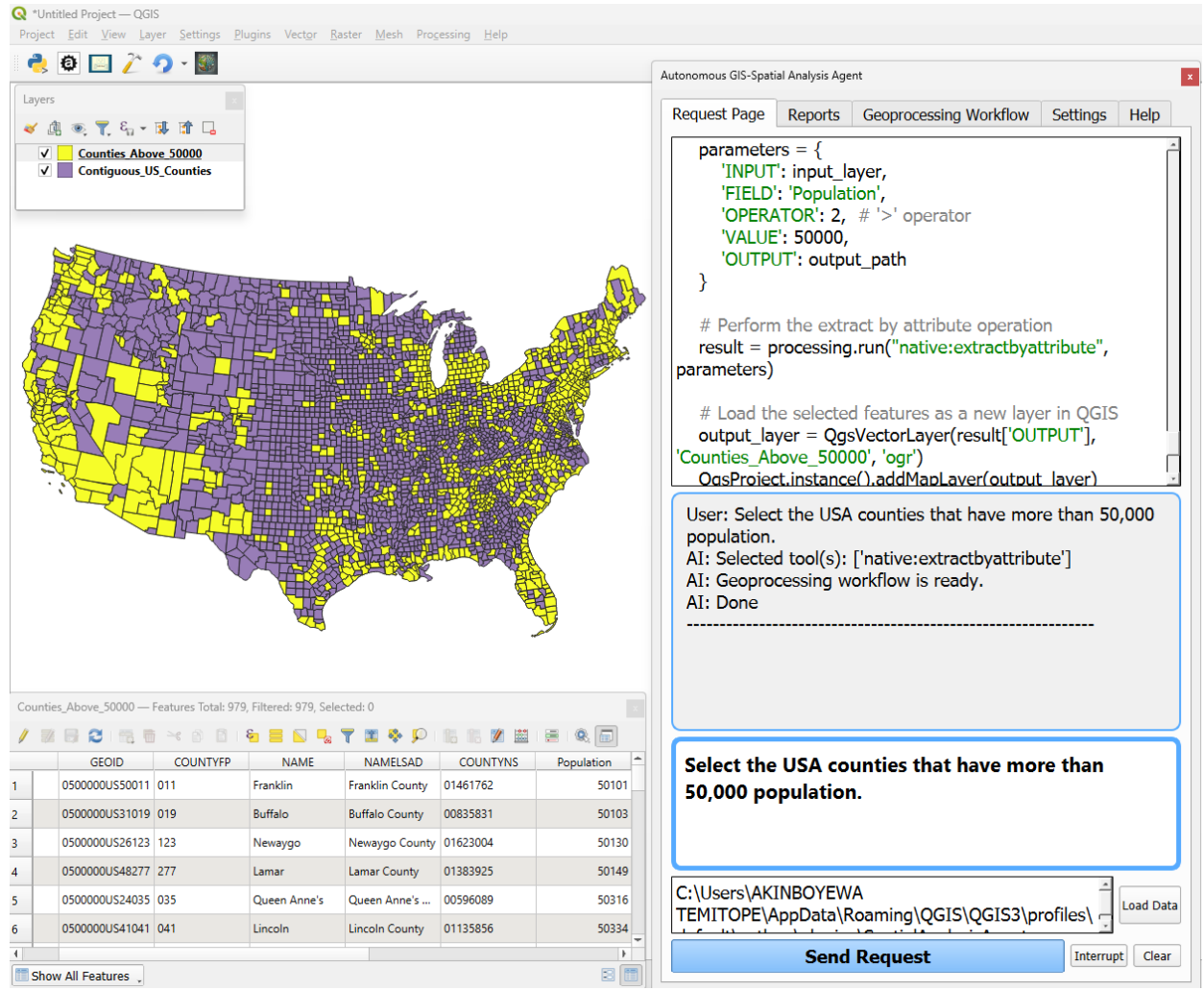

Figure 6: Selection of high population counties in the contiguous U.S. (Task: *Select the USA counties that have more than 50,000 population.*)

### 4.1.4 Extracting land cover data of Pennsylvania

This case focuses on extracting land cover data for Pennsylvania from a larger dataset covering the contiguous US. The input data includes land cover raster data from the National Land Cover Database (NLCD) and a shapefile containing the boundary of Pennsylvania. The agent accurately selected the appropriate tool (the GDAL Clip Raster by Mask Layer tool) and generated the executable code, configuring parameters such as the input layer, output path, and clipping settings. The agent was able to produce the clipped raster layer showing the land cover specific to Pennsylvania (Figure 7). The result provides a detailed view of various land cover categories within the state's boundary.

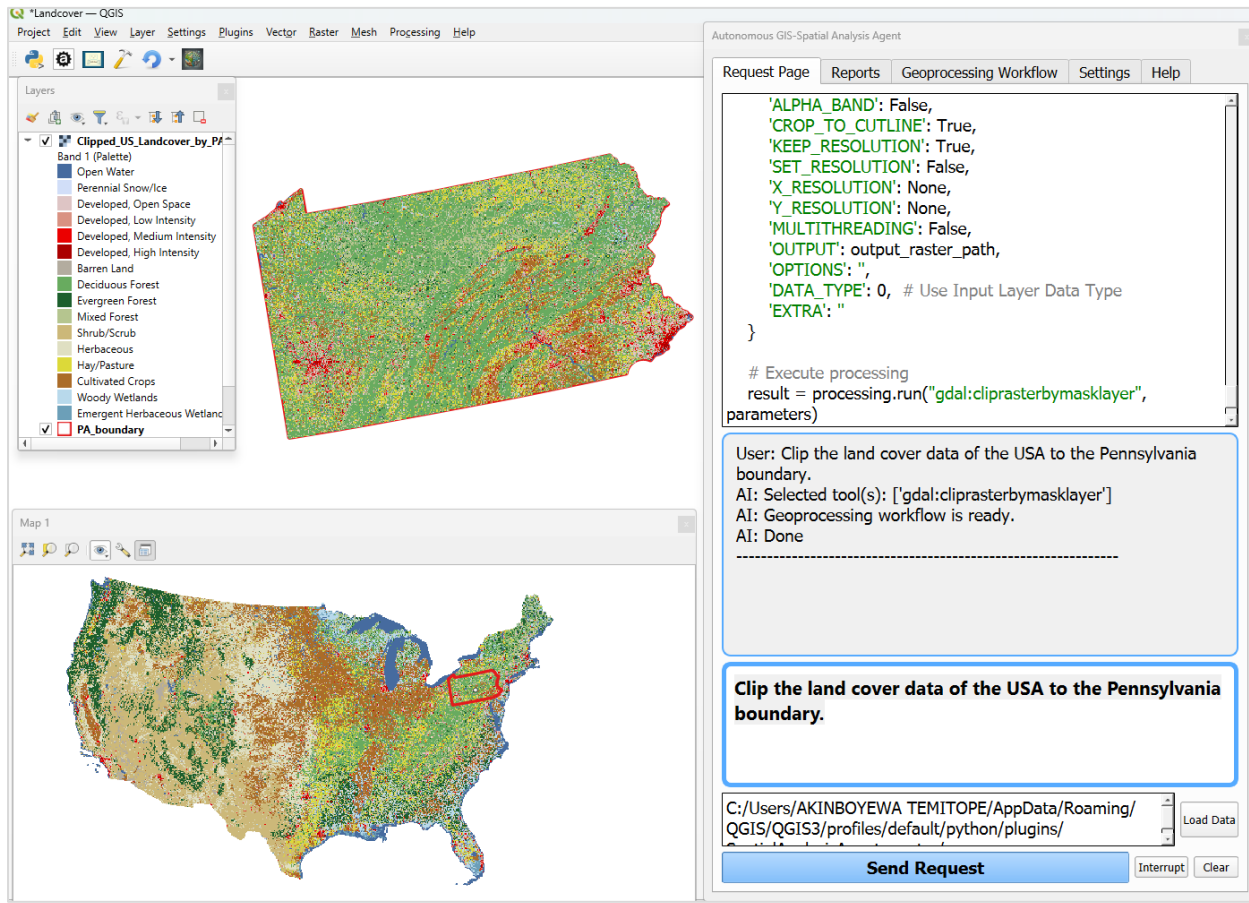

Figure 7: Extracting land cover information of Pennsylvania. (Task: *Clip the land cover data of the USA to the Pennsylvania boundary*.)

## 4.2 Intermediate level tasks

At this level, tasks become more complex and involve multiple steps and tools. The agent is guided with specific instructions or a list of steps to perform the task. Although the steps are outlined, the agent still needs to generate the correct code for each step and link them together to perform the analysis. The focus here is to test the agent's ability to follow a given general workflow while managing more than one process.

### 4.2.1 Zonal statistics of average elevation for counties in South Carolina

In this case, the agent is tasked with analyzing the average elevation for each county in South Carolina (SC) based on the DEM. This involved a series of geoprocessing steps: clipping the DEM to the boundary of SC, generating a histogram of the clipped DEM's pixel values, calculating zonal statistics, and creating a choropleth map. The input consists of three data layers: a DEM covering a region larger than SC and two vector layers containing SC's boundary and all SC counties.

The results of the request are shown in Figure 8. The choropleth map generated by the agent clearly displays the average elevation values, with darker shades representing higher elevations. Also, the histogram generated by the agent illustrates the distribution of pixel values from the clipped DEM. Additionally, the geoprocessing workflow diagram illustrates the steps of the analysis.

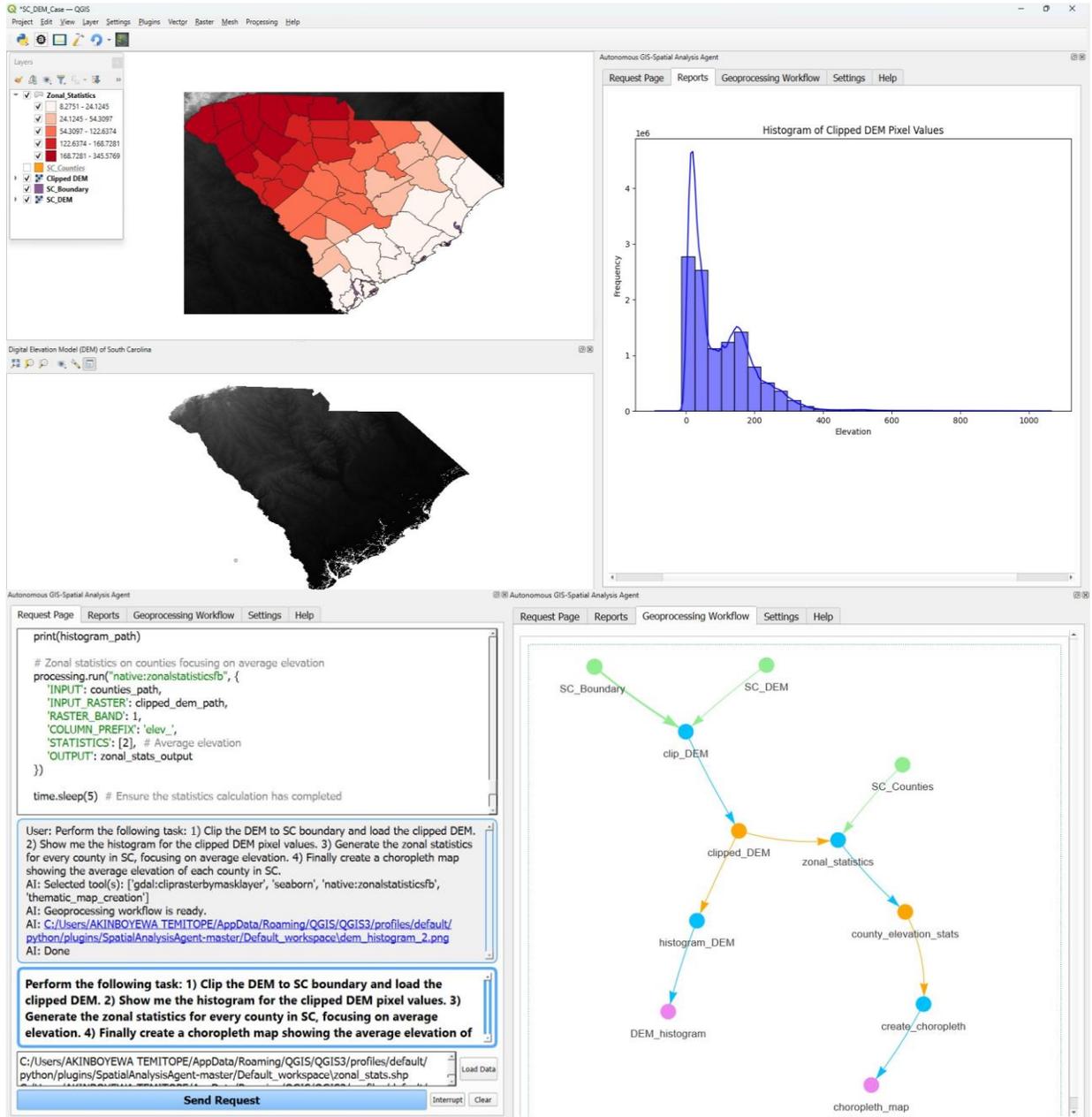

Figure 8: Zonal Statistics of average elevation for counties in South Carolina (*Task: Perform the following task: 1) Clip the DEM to SC boundary and load the clipped DEM. 2) Show me the histogram for the clipped DEM pixel values. 3) Generate the zonal statistics for every county in SC, focusing on average elevation. 4) Finally create a choropleth map showing the average elevation of each county in SC.*)

### 4.2.2 Richland County terrain analysis

In this case, the agent was tasked with performing a detailed terrain analysis for Richland County, SC, by merging four DEM tiles obtained from the US Geological Survey (USGS) and Shuttle Radar Topography Mission (SRTM) dataset (USGS, 2024). The generated geoprocessing workflow shows the sequence of steps to complete the task. The workflow began by merging the four individual DEMs to create an elevation model covering the area of interest. Once merged, the agent calculated multiple terrain characteristics, including slope, aspect, hillshade, Terrain Ruggedness Index (TRI), and Topographic Position Index (TPI). The outputs of each analysis step were displayed, with each terrain attribute as a separate map (Figure 9).

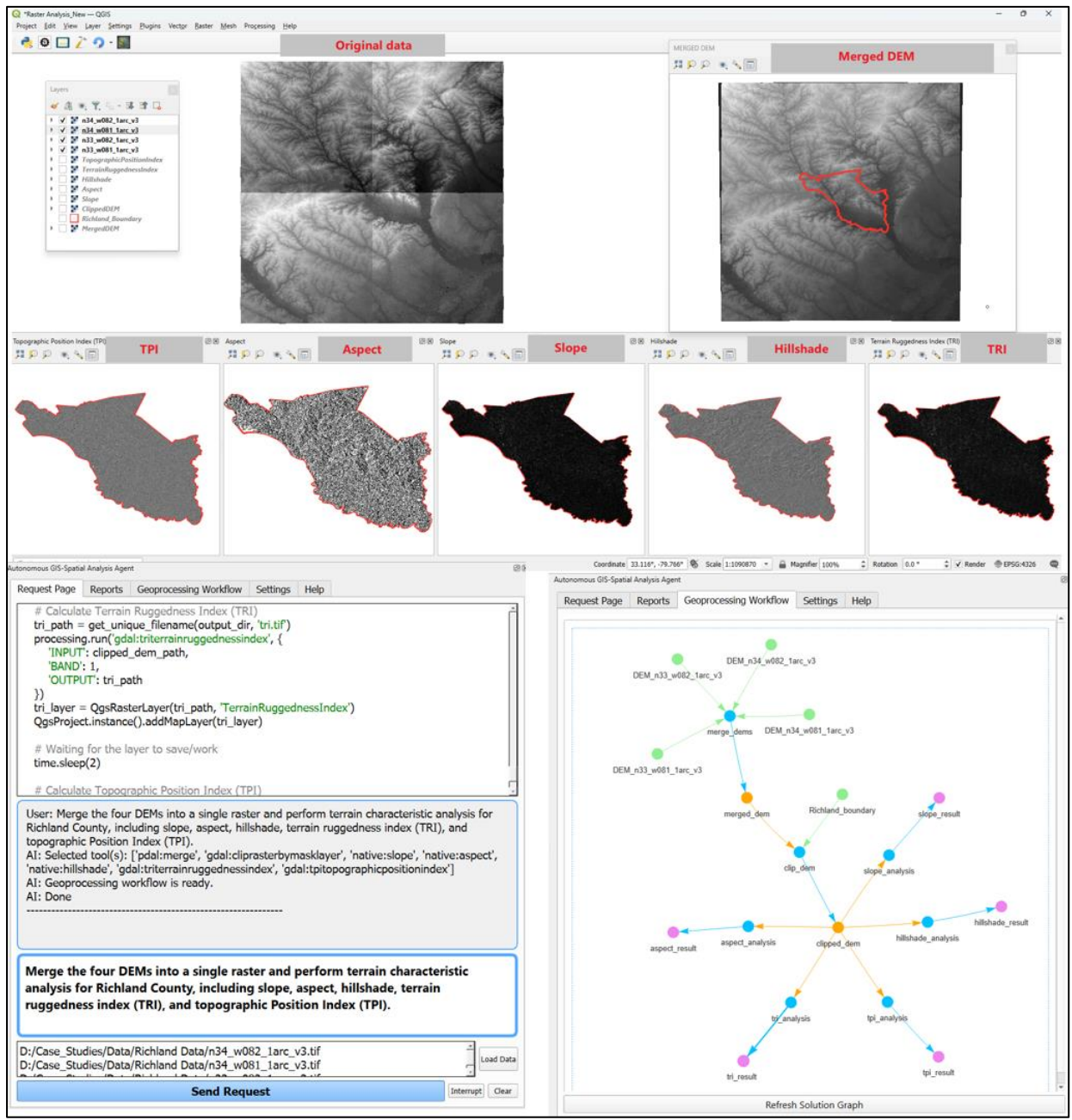

Figure 9: Richland County raster analysis. (Task: *Merge the four DEMs into a single raster and perform terrain characteristic analysis for Richland County, including slope, aspect, hillshade, terrain ruggedness index (TRI), and topographic Position Index (TPI).*)

### 4.2.3   County-level obesity risk behavior index analysis

This task involves the calculation of an obesity risk behavior index across all counties in the contiguous US. Two data layers – a shapefile containing the boundaries of all counties and a CSV file containing the rate of visits to different places such as convenience stores, limited-service restaurants, sport centers, fitness centers, and parks. The agent was able to join the attributes to the shapefile and select the fields needed for the analysis without any instruction from the user using the data understanding module. The thematic map and the geoprocessing workflow generated by the agent is presented in Figure 10.

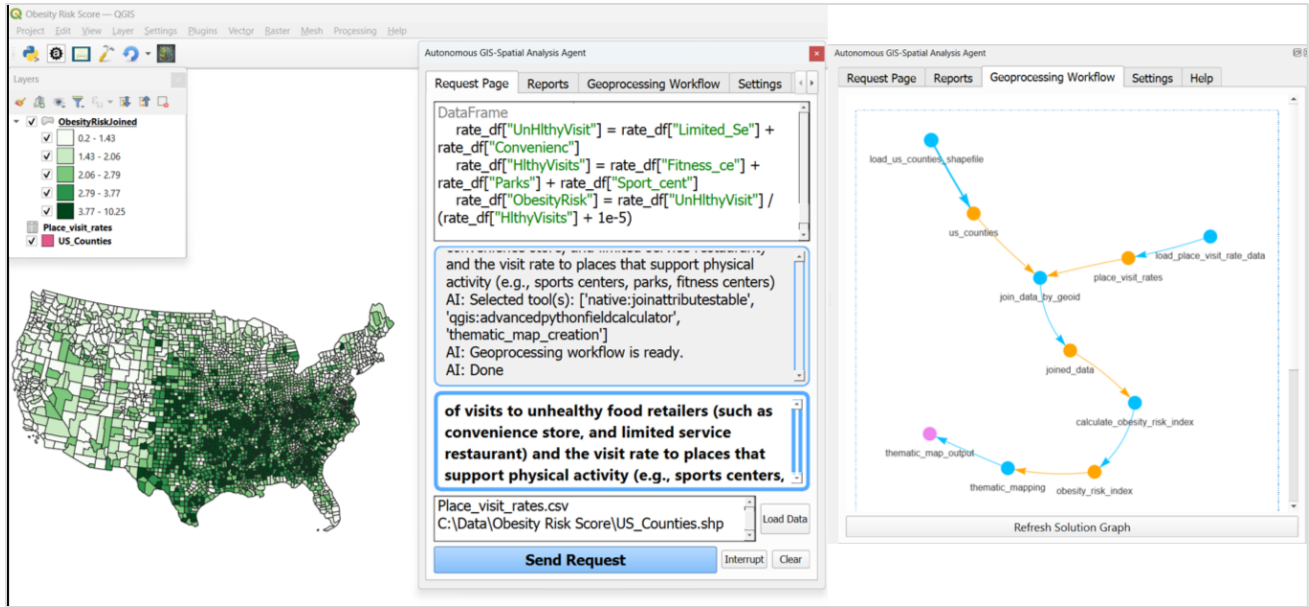

Figure 10: County-level obesity risk behavior index analysis (Task: *Generate an obesity risk behavior index of each county in the contiguous US by analyzing the rate of visits to unhealthy food retailers (such as convenience store, alcoholic drinking places, and limited service restaurant) and the visit rate to places that support physical activity (e.g., sports centers, parks, fitness centers). Visualize the results in a thematic map to highlight the obesity risk behavior index across counties.*).

### 4.3 Advanced level tasks

In the advanced level, tasks are multistep, and the agent is expected to determine the appropriate steps independently without explicit instructions, to devise the best approach for achieving the desired outcome. In other words, tasks are more natural in nature presenting the agent of "what" the users want with limited guidance on "how" to do it. The agent must select the right tools, generate the necessary code, and execute the entire process independently.

#### 4.3.1 Fast-food accessibility and obesity correlation analysis for Pennsylvania counties

This operation involves analyzing and visualizing the fast-food accessibility score for each county in Pennsylvania and performing a correlation analysis between the fast-food accessibility score and the prevalence of obesity. The agent's workflow began by calculating the fast-food accessibility score for each county based on the number of fast-food restaurants per capita by performing a spatial join operation. This was followed by generating a thematic map that displays counties with higher accessibility scores in darker shades of blue. Next, the agent analyzed the correlation between county-level obesity rates and fast-food accessibility scores. The results of this analysis are shown in Figure 11, where a scatter plot with a regression line highlights the relationship between fast-food accessibility and obesity rates across the state. The agent successfully managed the multistep process, calculating accessibility scores to perform correlation analysis, and visualizing the results in both a thematic map and a scatter plot. It should be noted that all the specific fields used in the analysis were automatically selected by the agent without explicit user guidance enabled by the data understanding module.

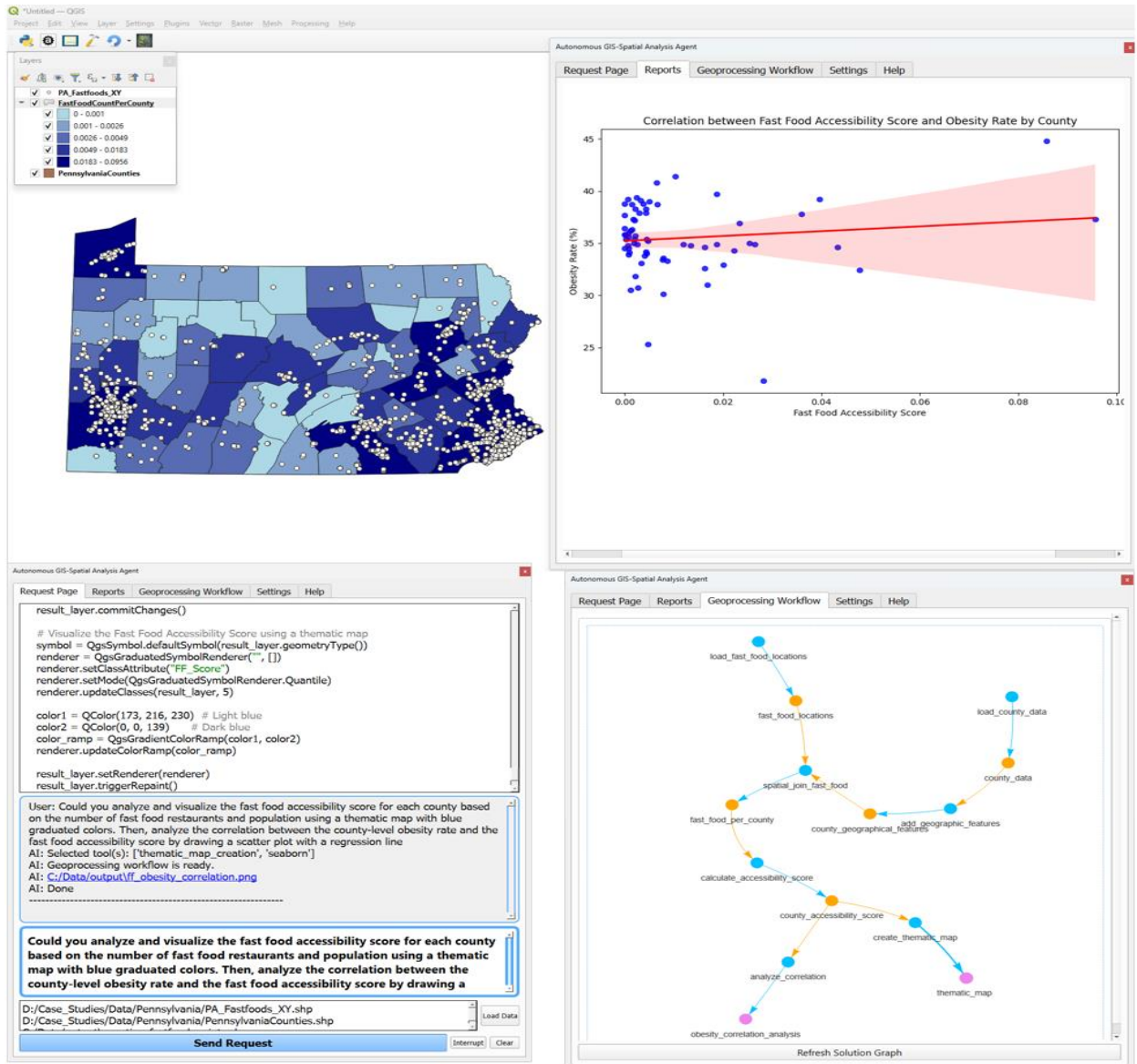

Figure 11: Fast-food accessibility analysis for each county in Pennsylvania. (Task: *Could you analyze and visualize the fast food accessibility score for each county based on the number of fast food restaurants and population using a thematic map with blue graduated colors. Then, analyze the correlation between the county-level obesity rate and the fast food accessibility score by drawing a scatter plot with a regression line.).* Fast food restaurants data was downloaded from OpenStreetMap. The population data (2022) was retrieved from the US census.

### 4.3.2 Spatial distribution of COVID-19 cases across US counties

This task asks, "Could you show the spatial distribution of the COVID-19 cases across US counties?" This request involves visualizing the spatial distribution of COVID-19 cases across US counties. The agent was provided with two datasets: shapefile of US county boundaries obtained from the US Census Bureau (2024) and a CSV file containing COVID-19 data as of December 2020, obtained from the New York Times (2023). The analysis involved joining the attribute contained in the CSV data with the shapefile to create a thematic map that shows the COVID-19 cases across all the counties in the US. The agent successfully completed the task, automatically identifying the common fields, "GEOID" and "FIPS", for the join operation. It also selected the "cases" field from the COVID-19 dataset to generate the thematic map. The geoprocessing workflow diagram generated by the agent and the resulting thematic map is shown in Figure 12. Counties with higher numbers of COVID-19 cases are represented by darker shades of blue on the map.

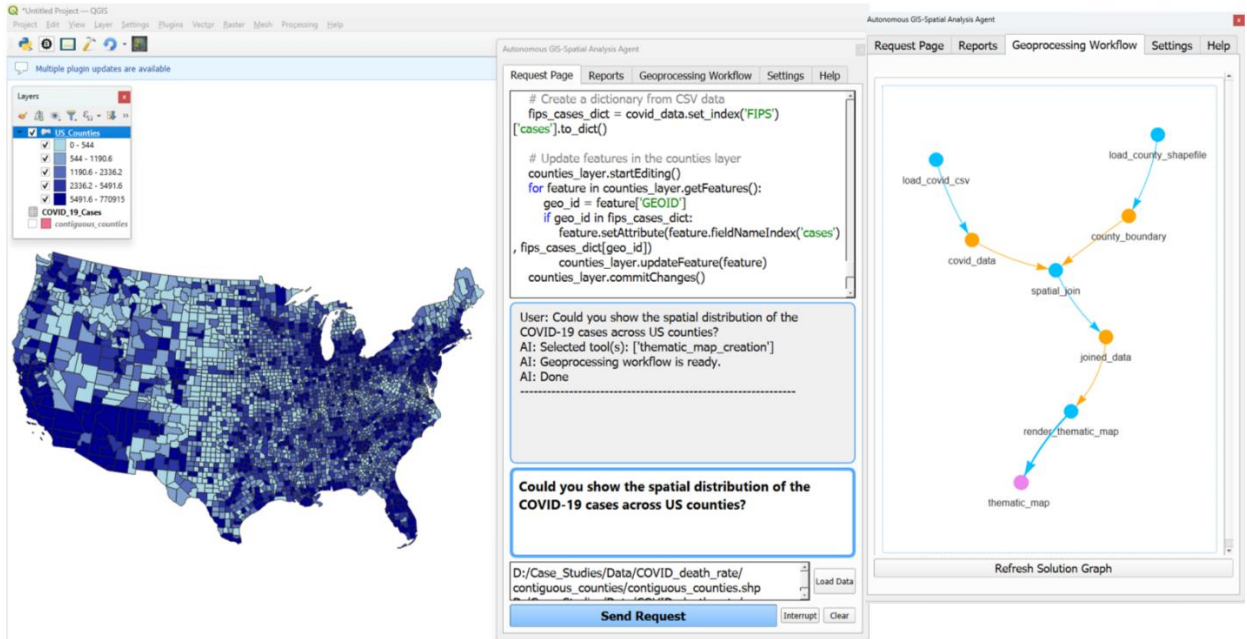

Figure 12: Spatial distribution of COVID-19 cases in the USA. (Task: *Could you show the spatial distribution of the COVID-19 cases across US counties?*)

### 4.3.3 Generating the Normalized Difference Vegetation Index (NDVI) from satellite imageries

In this case, the agent was tasked with generating the Normalized Difference Vegetation Index (NDVI) for the city of Akure, Nigeria using multispectral satellite imagery. The request is *"Generate the Normalized Difference Vegetation Index (NDVI) of Akure from these satellite imageries"*. The input data comprises a shapefile of the Akure boundary and Landsat 8 satellite products from the USGS, which includes multiple image bands. The agent demonstrated its capability by automatically identifying and selecting the correct bands, specifically the NIR and red bands, from the multispectral imagery. It then used these bands and applied the correct formula to generate the NDVI layer. The output was visualized as a raster map. The agent successfully managed the entire process, from band selection to calculation to ensure the correct bands were used and that the NDVI was calculated accurately (Figure 13). Note that the initial raster map generated by the agent used a black-and-white color ramp for symbology. The symbology was manually adjusted to enhance visualization in order to provide a clearer representation of the vegetation distribution in Akure, where areas of dense vegetation are represented by higher NDVI values, as shown in green, while shades of blue represent lower

NDVI values. This example shows the agent's ability to handle satellite imagery data and perform key calculations with minimal user input.

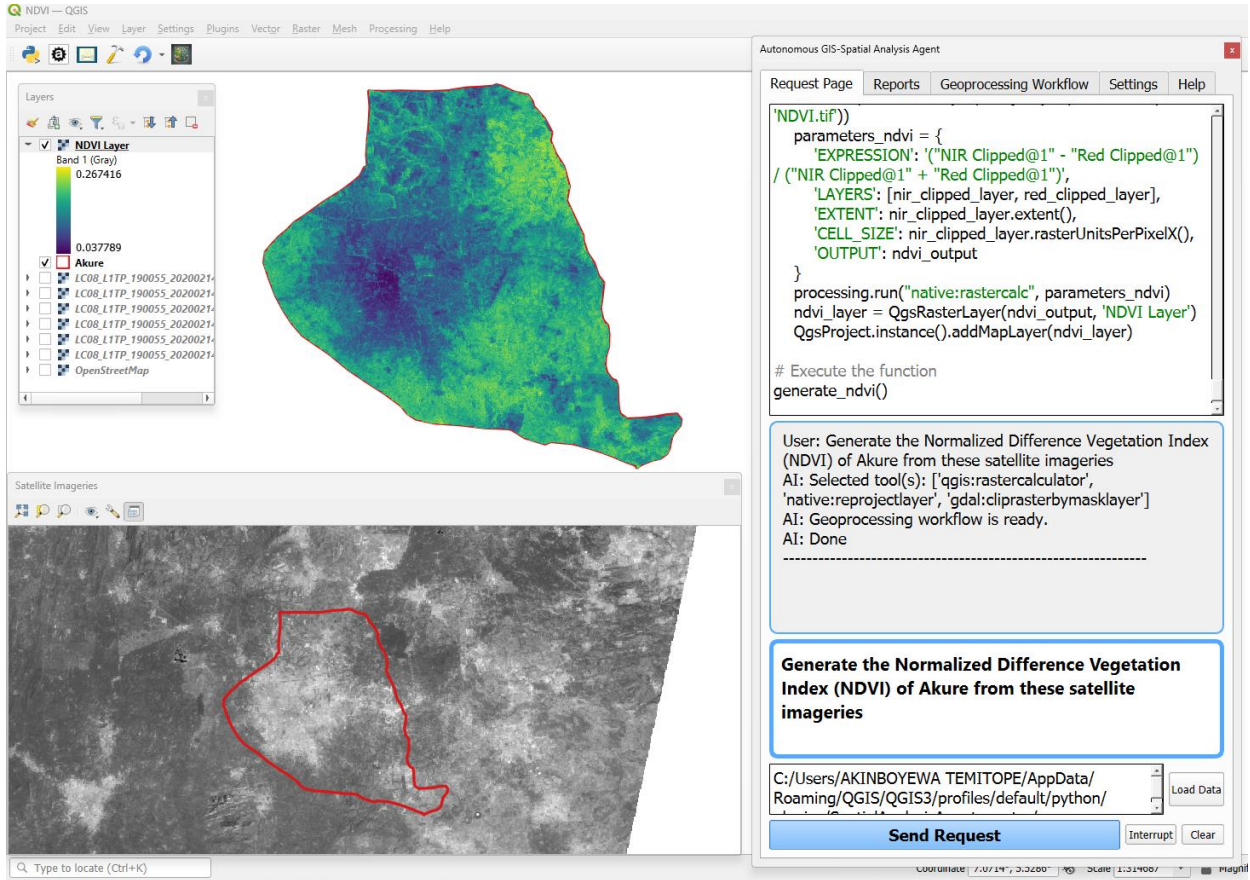

Figure 13: Generating the Normalized Difference Vegetation Index (NDVI) of Akure. (Task: *Generate the Normalized Difference Vegetation Index (NDVI) of Akure from these satellite imageries*.).

**4.4 Performance evaluation**

The performance of the GIS Copilot was assessed systematically using 110 tasks across three levels of complexity: basic, intermediate, and complex. Each level consisted of a set number of tasks, selected according to predefined criteria for task complexity (in Sections 4.1, 4.2, and 4.3). This structured approach ensured that the chosen tasks represented each complexity level to provide a balanced and meaningful evaluation framework. While definitions of complexity may vary, the selected tasks align closely with the GIS Copilot's intended functionality and scope.

Success rates were recorded at each stage, tracking tasks completed on the first attempt as well as those required multiple attempts. Tasks that remained unsuccessful after three attempts were categorized as failures. The overall accuracy is determined by calculating the percentage of successful cases. As shown in Table 1, with 60 cases at the basic level (Appendix 3), the success rate recorded was 95%; among 30 intermediate-level tasks (Appendix 4), a success rate of 80% was observed, and for the advanced level, which included 20 cases (Appendix 5), the success rate was 75%.

Table 1: Performance evaluation of the GIS Copilot

| Level of complexities | Total count of tasks | Count of successful cases | | | Count of failed cases | Overall success rate |
|---|---|---|---|---|---|---|
| | | First attempt | Second attempt | Third attempt | | |
| Basic | 60 | 52 | 3 | - | 5 | 92 % |
| Intermediate | 30 | 13 | 4 | 8 | 5 | 83 % |
| Advanced | 20 | 10 | 4 | 1 | 5 | 75 % |

We further evaluated the performance by analyzing the reasons behind the failed cases to understand the underlying issues. Overall, tool selection was accurate, as the appropriate tools were consistently chosen for each task. However, most of the failed cases are attributed to the following three major issues described below.

- **Incorrect parameter assignment**: The most frequent cause of failure involved parameter setting errors. Despite selecting the correct tool, some tasks failed due to incompatible or incorrectly assigned parameters, leading to code execution issues. For example, when attempting to identify areas suitable for camping, the agent was instructed to select locations where the aspect of the area was greater than 100 and the distance to the closest road did not exceed 1000 meters (Case 20 in Appendix 5). To achieve this, the agent used the *gdal:proximity* tool to calculate distances to the nearest roads. However, the task failed because the tool requires a raster input, while the agent assigned a vector layer as the input data.

- **Field name mismatches and data compatibility**: Another common issue arose from mismatches in field names during data joins, particularly when joining CSV files with shapefiles. Shapefiles often truncate field names to 10 characters, causing discrepancies that the agent did not account for, leading to calculation errors. For example, when tasked with calculating the population living in block groups containing hazardous waste sites, the agent needed to join a CSV file containing population data to a shapefile representing block groups. However, the agent encountered issues because the shapefile truncated longer field names to 10 characters. This caused the field names in the shapefile to differ from those in the CSV, leading to a failed join and, consequently, leading to a "field name error".

- **Projection issue**: During testing, it was observed that the agent frequently omitted essential reprojection steps unless specifically instructed. This oversight led to misaligned spatial data, affecting the accuracy of the results, particularly in tasks that required consistent coordinate systems across layers. For instance, when tasked with clipping an image using the extent of a shapefile where the shapefile and the image were in different coordinate systems, the agent did not reproject the shapefile to match the coordinate system of the image, the resulting clip operation was misaligned, leading to an error (Case 27 in Appendix 4).

## 5   Discussion

This study proposes a novel framework for integrating generative AI (LLMs) into established GIS platforms, using QGIS as a case study. Based on this framework, we implemented a GIS Copilot through which GIS users interact with the software with natural language. A key feature of this GIS Copilot is the extensive range of tools included, covering a broad spectrum of geoprocessing functions such as vector and raster analyses. Additionally, the framework is designed to be extensible by allowing GIS users to add custom tools following a predefined template. This extensibility ensures that the Copilot remains adaptable to user needs and the continuous development of new GIS tools and analyses. This work aligns with the increasing recognition of the potential for AI-driven systems to automate and streamline geoprocessing workflows. By automating these processes, the framework simplifies spatial analysis, making GIS functionalities more accessible to both expert and non-expert users.

This GIS Copilot responds to the previous work of Li and Ning (2023), who envisioned the need for a copilot-style system in GIS to enhance automation and interaction with geospatial data. They demonstrated the potential for LLMs to autonomously generate geoprocessing workflows and emphasized the importance of integrating these capabilities directly into existing GIS software to unlock their full potential. By implementing a Copilot system within QGIS, this study illustrates how LLMs can be used to generate executable spatial analysis algorithms using existing well-developed GIS tools. A significant benefit of Copilot style integration with GIS platform is that users have direct interaction with the agent directly within the GIS environment, allowing them to load and select datasets, visualize results and select between manual operations and agent-driven workflows based on their preferences. This flexibility enables users to combine traditional manual GIS processes with the efficiency of AI-driven analysis, resulting in a hybrid approach. For instance, users can load datasets into QGIS and perform analyses, generate results by the agent and further perform analyses using either the Copilot or manual methods.

In evaluating the performance of the GIS Copilot, we analyzed its success rate for over 100 tasks across three levels of complexity: basic, intermediate, and advanced level. At the basic level, tasks involved straightforward operations on one or two data layers, such as spatial buffering, calculating areas or filtering features based on attributes. The agent demonstrated high accuracy in selecting the appropriate tools and generating the necessary code. The reliability of the GIS Copilot at this level highlights its competence in automating foundational tasks with minimal intervention, offering an important utility for speeding up routine geospatial operations. At the intermediate level, the tasks require multistep which are described for the agent. Here, the agent not only needed to correctly select tools but also generate and execute the code in sequence. In these scenarios, the agent showed strong performance in generating geoprocessing workflows based on the multistep task description and managing the dependencies between different steps. The success in these tasks suggests that the agent can effectively operate as an assistant to handle more involved geospatial workflows, thereby reducing the need for users to do step-by-step manual coding. The advanced level tasks involve the highest complexity, where the tasks are more natural in nature, presenting the agent of "what" the users want with limited guidance on "how" to do it. Here, the agent needs to understand and reason the question and determine the geoprocessing workflow autonomously. This required the agent to interpret the natural language instructions, decide on an appropriate series of actions, select the correct tools, and execute the entire process without explicit guidance. While the agent showed promise at this level, especially in the tool selection and code generation, the success rate for these complex tasks was lower than simpler ones.

Explainability and trustworthiness are two critical concerns in generative AI applications, including GIScience (Tlili et al., 2023; Kang et al.,2024). While LLMs can generate impressive responses, their "black box" nature makes it difficult for users to understand how specific outputs are produced. Without clear explanations of how outputs are generated, users may be reluctant to fully adopt these models, especially when they play a significant role in data driven analyses and recommendations. To reduce these concerns, this GIS Copilot has been designed to enhance transparency at every step. Unlike traditional black-box models, the Copilot clearly presents each stage of the process, including tool selection and code generation, with generated code displayed in real time. Additionally, the generated code can be further edited and executed manually as needed. This approach allows users to inspect and understand each action taken by the Copilot, so that they can trace how results are achieved. Users are encouraged to review the generated code and results actively, rather than relying on them blindly. This approach highlights the collaborative nature of a "copilot" system that supports rather than replaces expert decision-making.

## 6 Limitations and Future Work

The GIS Copilot shows promising performance across the three levels of task complexities; however, certain issues were observed. These challenges point to potential areas for improvement and future research directions.

### 6.1 The need for a data validation module to reduce tool and parameter mismatch

The GIS Copilot shows potential in handling tasks of varying complexity, but issues like incorrect parameter assignment were observed, hindering its performance. For instance, the agent occasionally assigns vector data to parameters that require raster input. A good example of such is the *gdal:proximity* tool, where the input layer should be raster; however, the agent assigns a vector layer provided by the user without first converting it to raster, resulting in code failure. Improving the alignment between data types and tool requirements is needed to enhance the agent's reliability. One way this can be achieved is by integrating a *data validation module* that crosses-references each parameter with the data type required by the selected tool so that the data will be converted automatically even if there is a mismatch.

### 6.2 Expanding LLMs knowledge in GIScience fundamentals

LLMs like GPT-4o, while capable of general-purpose reasoning, exhibit limited foundational knowledge in GIScience, particularly in domain-specific requirements such as the need for data reprojection under certain conditions. During the advanced level testing, it was observed that the agent frequently omitted reprojection steps unless specifically instructed. This limitation affects the usability of the Copilot, especially for non-expert users who may be unaware of when or why reprojection is needed. These users may, therefore, unknowingly produce analyses based on misaligned or incorrectly scaled spatial data, which can lead to incorrect interpretations and reduced confidence in the results. Additionally, the agent demonstrated a lack of understanding of some specific GIS tools, especially some GDAL tools. Even with the integration of these tools' documentation in the framework, the agent still struggles to select appropriate functions or set accurate parameters for some tools, leading to execution errors. This may be due to limitations in prompt length. Some tool documentation lengths may exceed the standard prompt limit. To address these issues, future work could explore approaches like Retrieval-Augmented Generation (RAG) (Lewis et al., 2020), which may address the length constraints within tool documentation, by allowing the LLM to query and retrieve specific and relevant sections of documentation as needed.

### 6.3 Building an experience-based knowledge system

The agent's current knowledge is mostly derived from Internet sources, which lack the depth of practical GIS experience necessary for accurately performing specialized tasks. For instance, the agent may not recognize that data types in table joins must align across columns, such as ensuring that strings match with strings and integers with integers, or that buffer distances should be expressed in metric units rather than degrees. Without this foundational understanding, the agent is prone to errors that are easily detectable by GIS professionals yet may remain undetectable by less or non-experienced users. To address this issue, implementing an *experience-based knowledge system* could improve the agent's practical understanding and adaptability. Such a system would act as a database that accumulates instances of errors and their solutions during agent-user interactions. Over time, this "experience database" would function similarly to a repository of case-based learning, enabling the agent to recognize recurring issues, learn corrective actions, and apply these solutions autonomously in future tasks. This approach is one way of achieving the "self-growing" goal of an autonomous GIS, allowing the agent to continuously evolve and improve based on accumulated experience. For example, if the agent encounters a table join error due to incompatible data types, it will reference its experience database to identify and implement a solution, such as converting one of the columns to the appropriate type.

### 6.4 Integration of a prompt-generation module

In this study, spatial tasks were provided directly to the agent as prompts. However, these task descriptions are not necessarily the appropriate prompt as they may often lack the structure and detail required, especially in technical contexts. For instance, a simple task description may omit key parameters or methodological details that might help the agent to perform more accurately. To address this, one potential approach is to introduce an intermediate step that refines or rephrases task descriptions into optimized prompts. This step could involve analyzing the initial task description, extracting essential elements such as data type specifications, required tool parameters, and expected outcomes, and transforming them into a more detailed, structured prompt specific to different types of geospatial tasks. By providing a richer context, the refined prompt would guide the agent in selecting appropriate tools and parameters to ensure greater alignment with the intended analysis. For instance, it could recognize when a task requires specific spatial projections and then structure the prompt appropriately.

### 6.5 Integration with a geospatial data retrieval agent

To advance the GIS Copilot toward full autonomy, integrating an agent capable of retrieving needed geospatial data is essential. An autonomous data retrieval agent would allow the Copilot to access and use relevant datasets directly, thereby reducing user input requirements and improving efficiency, especially for non-experts who may lack familiarity with data sourcing. Recently, Ning et al. (2024) introduced an agent for retrieving geospatial data from different sources, implemented as a QGIS plugin. Integrating such an agent could help the Copilot access to the data needed for analysis without requiring the user to supply it, which could bring the GIS Copilot closer to a fully spatial analysis agent.

### 6.6 Exploring the integration of open-source LLM models

In this study, OpenAI GPT has been used as the primary language model to power the GIS Copilot to demonstrate the potential of the proposed approach. However, integrating open-source LLM models such as Meta's LlaMA (Meta, 2024) could offer notable advantages, especially in terms of customization and accessibility, such as fine-tuning on GIS-specific tasks to further improve performance and alignment with specialized geospatial needs. Future studies could explore the integration of these open-source models into the GIS Copilot to provide a more accessible and customizable alternative to commercial models.

## 7 Conclusions

The integration of generative AI into GIS represents a significant advancement in automating spatial analysis tasks. This study presents a framework for integrating LLMs directly into existing GIS platforms. The framework was implemented as a GIS Copilot in the QGIS environment. Through case studies across different levels of task complexity, the GIS Copilot proved effective in automating basic and intermediate geospatial tasks, reducing the need for manual operations and coding. While the Copilot performed well in these simpler tasks, challenges remain in achieving full autonomy for more advanced and multistep processes without explicitly telling it "how".

Despite the limitations, the GIS Copilot represents a promising step toward the development of fully autonomous GIS and opens avenues for future exploration of AI-driven GIS solutions in GIScience. This study contributes to the broader vision of autonomous GIS, where AI-powered agents can autonomously perform spatial analysis to democratize access to geospatial analysis, transforming how users engage with geospatial data and spatial problems. As advancements in generative AI capabilities continue, further research will focus on improving the Copilot's performance in handling complex workflows, enhancing its error-correction mechanisms, and expanding its adaptability to other GIS platforms.

**Data and code availability statement:** The source code for the GIS Copilot is available on GitHub at https://shorturl.at/vRcm6. The GIS Copilot can be downloaded from the official QGIS plugin page at https://plugins.qgis.org/plugins/SpatialAnalysisAgent-master and installed by following the instructions provided in the GIS Copilot User Manual. The datasets used for testing alongside with the testing cases can be accessed through https://shorturl.at/bI4Ep.

# Appendices

## Appendix 1: Sample of QGIS native tool documentation (structured in TOML format)

```
tool_ID = 'native:clip'

tool_name = 'Clip'

brief_description = ''' Clips a vector layer using the features of an additional polygon layer.
'''

parameters ='''
INPUT: Layer containing the features to be clipped

OVERLAY: Layer containing the clipping features

OUTPUT: Specify the output (buffer) layer. One of: Create Temporary Layer (TEMPORARY_OUTPUT); Save to File…; Save to Geopackage…; Save to Database Table…; Append to Layer…
'''

code_example = '''
import processing
from qgis.core import QgsProject,QgsVectorLayer
def clip():
    output_path ="D:/Output_path/Output_layer.shp"

    # Define the parameters Example below:
    parameters = {
        'INPUT': 'D:/SpatialAnalysisAgent/Data/BG.shp',
        'OVERLAY': 'D:/SpatialAnalysisAgent/Data/bd.shp',
        'OUTPUT':output_path}
# Perform the extract by attribute operation
    result = processing.run("native:clip", parameters)
    # Load the selected features as a new layer
    output_layer = QgsVectorLayer(result['OUTPUT'], 'Clipped Layer', 'ogr')
    QgsProject.instance().addMapLayer(output_layer)
clip()
'''
```

```
# Load the selected features as a new layer
   output_layer = QgsVectorLayer(result['OUTPUT'], 'Clipped Layer', 'ogr')
   QgsProject.instance().addMapLayer(output_layer)
clip()
'''
```

**Appendix 2: Sample of customized tool documentation (structured in TOML format)**

```
tool_ID = 'densityMap_kerneldensityestimation'
tool_name = 'Heatmap (Kernel Density Estimation) '
brief_description = '''
 Creates a density (heatmap) raster of an input point vector layer using kernel density estimation. Heatmaps allow easy identification of hotspots and clustering of points.
   The density is calculated based on the number of points in a location, with larger numbers of clustered points resulting in larger values.
'''
parameters ='''
code_example = '''
import geopandas as gpd
import numpy as np
from scipy.stats import gaussian_kde
import rasterio
from rasterio.transform import from_origin
# Load the shapefile and ensure it's in a projected coordinate system
gdf = gpd.read_file('D:/Case_Studies/Data/PA_restaurants/PA_fastfood_restaurants.shp').to_crs(epsg=26918)
# Extract coordinates for KDE
coords = np.vstack([gdf.geometry.x, gdf.geometry.y])
kde = gaussian_kde(coords)
```

```python
# Define grid and calculate KDE density
xmin, ymin, xmax, ymax = gdf.total_bounds
xgrid = np.linspace(xmin, xmax, 200)
ygrid = np.linspace(ymin, ymax, 200)
x, y = np.meshgrid(xgrid, ygrid)
density = kde(np.vstack([x.ravel(), y.ravel()])).reshape(x.shape) * 1e6  # scaled density

# Export the KDE result as a GeoTIFF
output_file = r'C:\Users\Output\kde_output.tif'
transform = from_origin(xmin, ymax, xgrid[1] - xgrid[0], ygrid[1] - ygrid[0])
with rasterio.open(
    output_file, 'w',
    driver='GTiff', height=density.shape[0], width=density.shape[1],
    count=1, dtype=density.dtype, crs=gdf.crs, transform=transform
) as dst:
    dst.write(density, 1)

# Display the result in QGIS
output_layer = QgsRasterLayer(output_file, 'KDE_Output')
QgsProject.instance().addMapLayer(output_layer)
```

# Appendix 3: Basic level test cases

Note that all datasets included here are for testing purposes only.

| S/N | Task | Category | Tool(s) used | Number of attempts | Dataset used | Screenshot |
|---|---|---|---|---|---|---|
| 1 | Select the counties that have less than 10% increase in population from 2005 to 2008. | Vector Selection | native:joinattributestable | 1 | SC_Counties | Population_Increase2005_2008 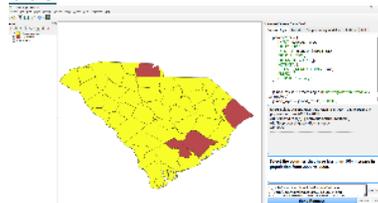 |
| 2 | Create barchart showing obesity prevalence in the United States. Ensure all labels are clear | Plot | seaborn | 1 | US_Counties | Obesity_Barchart 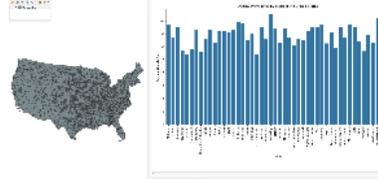 |
| 3 | Clip the DEM to PA Boundaries | Raster Analysis - Clipping | gdal:cliprasterbymasklayer | 1 | PA_DEM | PA_DEM_Clipping 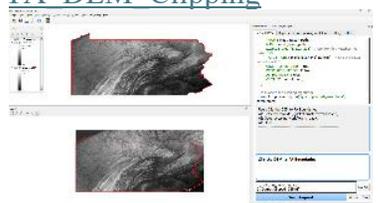 |
| 4 | Generate contour lines from the DEM of Puerto Rico with a 50-meter interval | Raster Analysis - Contour Generation | Gdal:contour | 1 | Puerto Rico DEM | Puerto_Rico_Contour 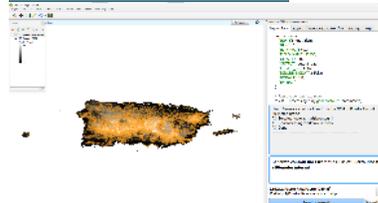 |

| S/N | Task | Category | Tool(s) used | Number of attempts | Dataset used | Screenshot |
|---|---|---|---|---|---|---|
| 5 | Add new fields for latitude and longitude to the fast food restaurants layer | Add Geometry Attributes | Native:addxyfields | 1 | PA_Fastfoods | Add XY_to_Fastfood 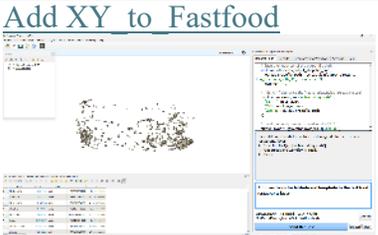 |
| 6 | Extract the counties with Median household below 50,000 USD in Pennsylvania. | Vector Selection - Attribute Query | qgis:selectbyexpression | 1 | US_Counties | PA_MedInc_below50000 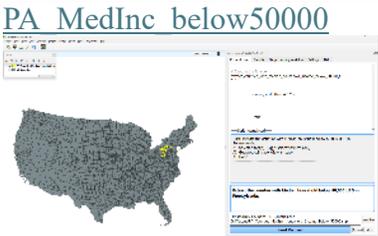 |
| 7 | How many damaged houses were located within the San Andreas Special Studies Zones? | Spatial Query | native:extractbylocation | 1 | DamagedHouses SAF_SpecialStudyZone.shp | DamagedHouses 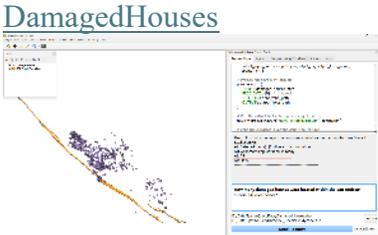 |
| 8 | Generate point features at the intersections of railways and the road features | Vector Overlay Analysis | native:lineintersections | 1 | Nigeria_Major_Roads Nigeria_Railways | Railway_Road_Intersection 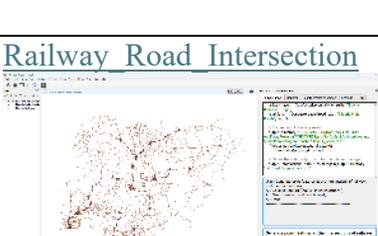 |
| 9 | Apply an affine transformation to the | Geometric Transformation | native:affinetransform | 1 | Penn_State_Building | Affine transform 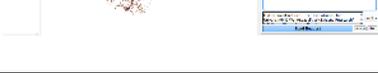 |

| S/N | Task | Category | Tool(s) used | Number of attempts | Dataset used | Screenshot |
|-----|------|----------|--------------|-------------------|--------------|------------|
|  | building layer to scale by a factor of 2 and rotate 30 degrees. |  |  |  |  | 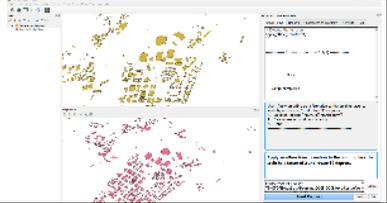 |
| 10 | Based on the provided data layers, draw a scatter plot for the Obesity rate and Covid-19 death rate. | Statistical Analysis- plot | seaborn | 1 | US_Counties_Obesity_Covid | Covid_Obesity_ScatterPlot 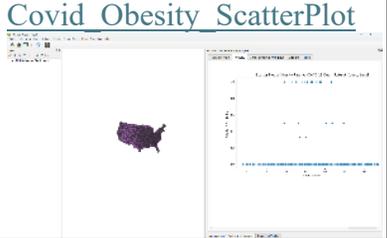 |
| 11 | Report the information for this raster data in HTML. | Raster Data Reporting | gdal:gdalinfo | 1 | SC_DEM | SC_Raster_Data_Report 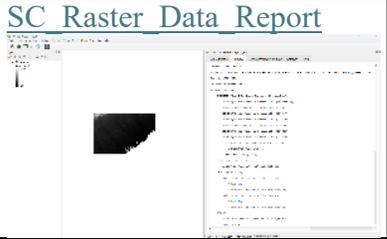 |
| 12 | Can you show me the histogram for the DEM pixel values? | Raster Analysis - histogram | seaborn | 1 | SC_DEM | SC_DEM_Histogram 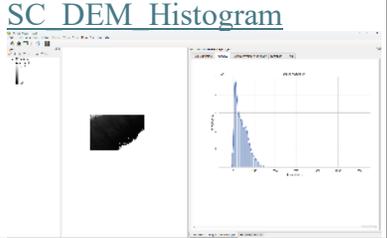 |
| 13 | Do a Fourier conversion for the loaded image. | Image processing | numpy | 1 | LandSat/LT05_L1TP_123042_20110928_20200820_02_T1_B5.TIF | Fourier1 |

| S/N | Task | Category | Tool(s) used | Number of attempts | Dataset used | Screenshot |
|---|---|---|---|---|---|---|
| | | | | | | 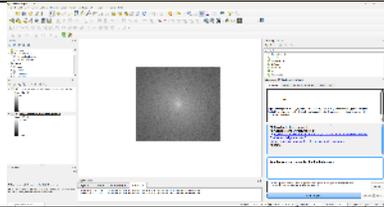 |
| 14 | Composite the three bands of LandSat 5 to a RGB image using the most commonly used band composition. | Image processing | gdal:rearrange_bands | 2 | LandSat/TM5_L2/LT05_L2SP_029030_20110605_20200822_02_T1_SR_B3.TIF  LandSat/TM5_L2/LT05_L2SP_029030_20110605_20200822_02_T1_SR_B2.TIF  LandSat/TM5_L2/LT05_L2SP_029030_20110605_20200822_02_T1_SR_B4.TIF | LandSat_band_composition_success 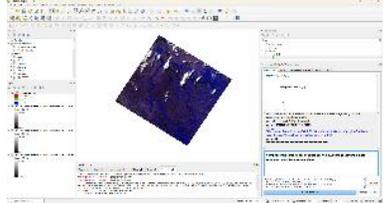 |
| 15 | Separate each band of the image into a single TIF file. | Image processing | gdal:translate | 1 | LandSat\TM5_L2\composited_rgb.tif | separate_bands 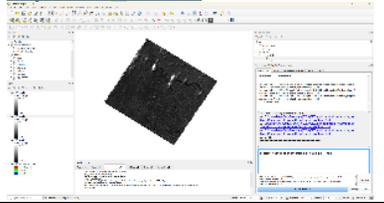 |
| 16 | Create a histogram of the distance from the flood event location to its nearest river. | Vector overlay | qgis:distancetonearesthubpoints, seaborn | Failed. | Flood2015may.shp  River.shp | |

| S/N | Task | Category | Tool(s) used | Number of attempts | Dataset used | Screenshot |
|---|---|---|---|---|---|---|
| 17 | Compute the distance of each flood to the nearest river. | Proximity Analysis | ['qgis:distancetonearesthubpoints'] ['native:hublines'] | Failed | rivers_ESRI102965.shp, flood2015may_ESRI102965.shp | |
| 18 | Draw the shortest line between the flood locations and rivers. | Distance Analysis | ['native:shortestline'] | 1 | rivers_ESRI102965.shp, flood2015may_ESRI102965.shp | 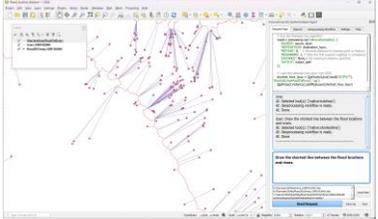 flood_nearest_river.png |
| 19 | Reproject the loaded layers to ESRI:102005 USA_Contiguous_Equidistant_Conic. | Reprojection | 'gdal:warpreproject' | 1 | Flood2015may.shp, River.shp | Projection 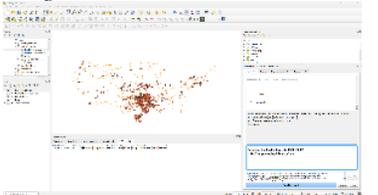 |
| 20 | Reproject all loaded layers into EPSG6569, and save them into the same folder, adding "EPSG6569" to the current basenames. | reprojection | ['native:reprojectlayer'] | 1 | Columbia_schools.shp, Columbia_sidewalk.shp, Columbia_roads | Snipaste_2024-11-04_18-09-38.png 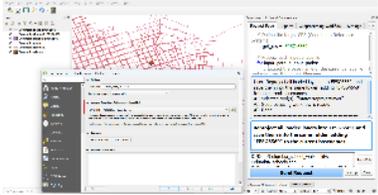 |
| 21 | I want you to convert this csv file into a shapefile. The longitude is presented as 'X' | Data Conversion | ['native:createpointslayerfromtable', | 1 | HIV.csv | Level1_25 |

| S/N | Task | Category | Tool(s) used | Number of attempts | Dataset used | Screenshot |
|---|---|---|---|---|---|---|
| | and the latitude is in column 'Y'. | | 'native:savefeatures', 'native:loadlayer'] | | | |
| 22 | I want you to only keep US contiguous counties from this input data. | Geographic Filtering | ['native:extractbyexpression'] | 2 | US counties shapefile | Level1_26 |
| 23 | I want you to calculate the length of each polyline and then add a layer named 'length'. | Geometric Calculation | ['qgis:exportaddgeometrycolumns'] | 1 | tl_2019_42_prisecroads | Length Calculation |
| 24 | I want you to buffer this polyline and the buffer distance should be 0.1m. | Buffering Analysis | ['native:reprojectlayer', 'native:buffer'] | 1 | Network_Analysis/network_points | Buffer |
| 25 | I want you to clip all counties of Pennsylvania from this input data. | Clip Analysis | ['native:extractbyattribute', 'pdal:clip'] | 1 | US counties | Level1_30 |

| S/N | Task | Category | Tool(s) used | Number of attempts | Dataset used | Screenshot |
|---|---|---|---|---|---|---|
| | | | | | | 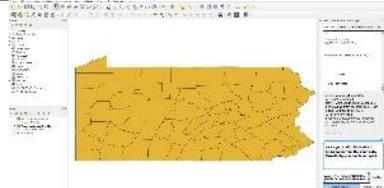 |
| 26 | I want you to identify all points that intersect with Pennsylvania _Counties_Clip data. | Spatial Intersection Analysis | ['native:reprojectlayer', 'native:extractbylocation'] | 1 | US counties  HIV csv file | Level1_31 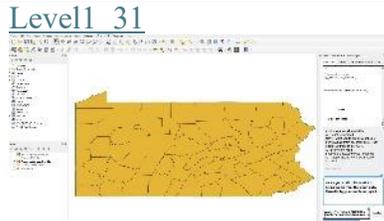 |
| 27 | I want you to find the difference of the two shapefiles that overlap. | Symmetrical Difference Analysis | ['native:difference'] | 1 | Generated a polygon, PA counties | Level1_32 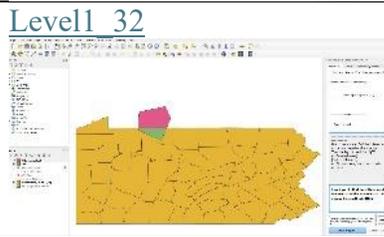 |
| 28 | I want you to dissolve the selected shapefile. | Dissolve Operation | ['native:dissolve'] | 1 | PA counties | Dissolve 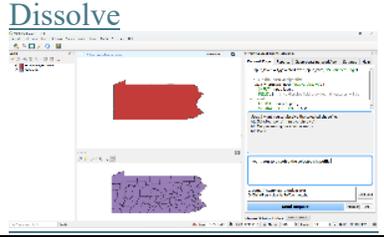 |
| 29 | I want you to apply symmetrical difference operations on these two selected shapefiles. | Symmetrical Difference Analysis | ['native:symmetricaldifference'] | 1 | PA counties and generated polygon | Level1_34 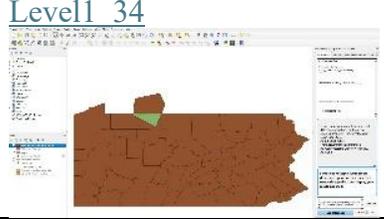 |

| S/N | Task | Category | Tool(s) used | Number of attempts | Dataset used | Screenshot |
|---|---|---|---|---|---|---|
| 30 | I want you to apply union operations on these two selected shapefiles. . | Union Analysis | ['native:union'] | 1 | PA counties and the generated polygon | Level1_35 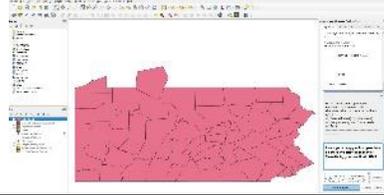 |
| 31 | I want you to apply eliminate selected polygons operations on these two selected shapefiles. | Polygon Elimination | ['qgis:eliminateselectedpolygons'] | 1 | PA counties and the generated polygon | Level1_36 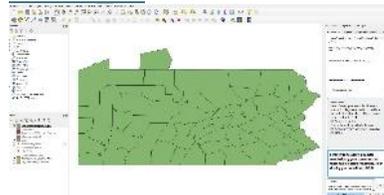 |
| 32 | I want you to generate the centriod of all these counties that are in this shapefile. | Centroid Calculation | ['native:centroids'] | 1 | PA counties | Level1_37 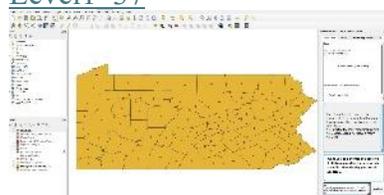 |
| 33 | Generate a shapefile from the CSV file. | Data Conversion | ['native:createpointslayerfromtable'] | 2 | Street_view_image\Richland_jsons.csv | CSV to shapefile 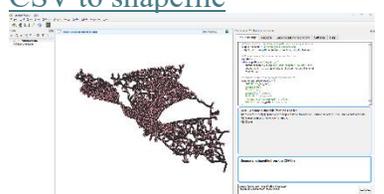 |
| 34 | Extract the boundary polygon of Laramie County and save it to the individual shape file. | Boundary Extraction | ['native:extractbyattribute'] | 1 | Wyoming_Counties_projected.shp | Extract_save_feature.png 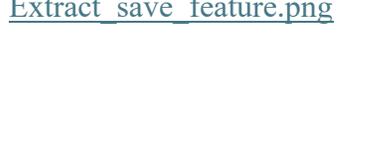 |

| S/N | Task | Category | Tool(s) used | Number of attempts | Dataset used | Screenshot |
|---|---|---|---|---|---|---|
|  |  |  |  |  |  | 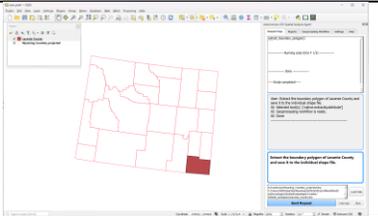 |
| 35 | From this input DEM data, I want you to generate a hillshade | Terrain Analysis | ['native:hillshade'] | 1 | n40_w078_1arc_v3 | Level1_40 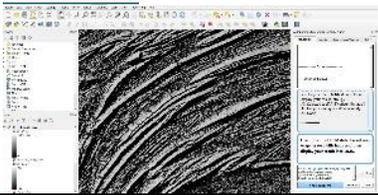 |
| 36 | Sum all the pixel values inside a Census tract polygon for each tract. | Zonal Statistics | ['native:zonalstatisticsfb'] | Failed | LandScan\ landscan-usa-2021-conus-night.tif, tract_37_EPSG4326 |  |
| 37 | Provide pixel statistics inside the polygon. | Zonal Statistics | ['native:zonalstatisticsfb'] | 1 | LandScan\Laramie.shp, LandScan/landscan-usa-2021-conus-night.tif | 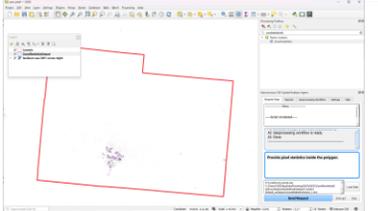 Statistic_pixel.png |

| S/N | Task | Category | Tool(s) used | Number of attempts | Dataset used | Screenshot |
|---|---|---|---|---|---|---|
| 38 | Reproject the DEM to EPSG6569, and set the cell size to 30 * 30 meters. | Raster Reprojection | ['gdal:warpreproject'] | 1 | Richland_DEM.tif | 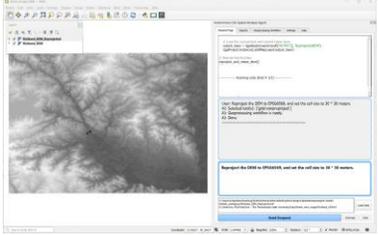 DEM_reproject.png |
| 39 | Assign EPSG6569 to the file. | Coordinate System Assignment | ['pdal:assignprojection'] | 1 | Richland_DEM_EPSG6569.tif | 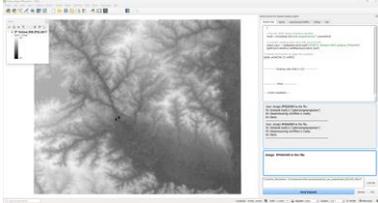 DEM_assign_project.png |
| 40 | Find out the number of fastfood restaurants in each county of Pennsylvania | Spatial Join Analysis | native:joinbylocationsummary | Failed | PennsylvaniaCounties PA_Fastfoods | |
| 41 | Generate 20 random points inside this polygon | Random Point Generation | ['qgis:randompointsinsidepolygons'] | 1 | Circle.shp | Level1_39 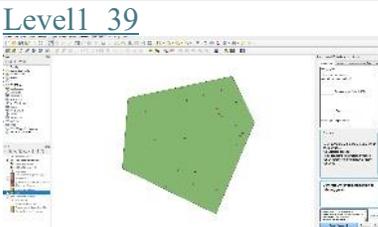 |
| 42 | Apply extract layer extent operation based on these two input shapefiles. | Extent Extraction | ['native:polygonfromlayerextent'] | 1 | Circle.shp | Level1_40_0 |

| S/N | Task | Category | Tool(s) used | Number of attempts | Dataset used | Screenshot |
|---|---|---|---|---|---|---|
| | | | | | | 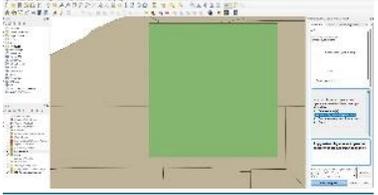 |
| 43 | Generate grid inside this selected polygon and the cell size should be 2 by 2 meters. | Grid Generation | ['native:creategrid'] | 1 | Rectangle_grid.shp | Level1_41 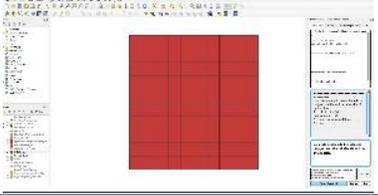 |
| 44 | Generate a distance matrix table showing distances between all DC health facilities. | Distance Matrix Calculation | qgis:distancematrix | 1 | DC health center/DC_Health_Facilities | Distance_Matrix 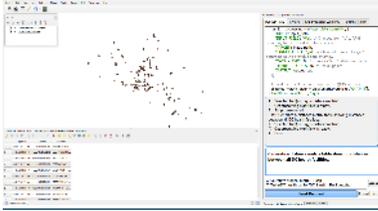 |
| 45 | Based on these two input shapefiles, please conduct mean coordinate operation. | Mean Center Calculation | ['native:meancoordinates'] | 1 | Point_inside_polygon.shp; Rectangle_grid.shp | Level1_43 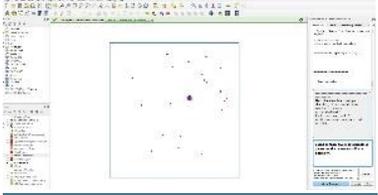 |
| 46 | Identify the 5 nearest neighbors to the center point from the points dataset. | Nearest Neighbor Analysis | ['native:joinbynearest'] | failed | Points.shp; Center.shp | |
| 47 | Generate regularly spaced points inside the selected | Point Grid Generation | ['qgis:regularpoints'] | 1 | Rectangle_grid.shp | Level1_46 |

| S/N | Task | Category | Tool(s) used | Number of attempts | Dataset used | Screenshot |
|---|---|---|---|---|---|---|
| | polygon, ensuring that the distance between each point is 0.1 meters. | | | | | 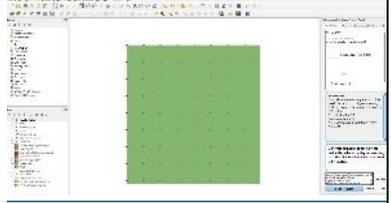 |
| 48 | Using raster calculator, convert the nighttime satellite image pixel digital number (DN) to radiance (L) using the formula: $L=DN^{(1.5)}*10e-10$ | Raster calculation. | ['qgis:rastercalculator'] | 1 | Nighttime/ LuoJnia1-01_LR201903142986_20190311145630_HDR_0000_gec.tif | Nighttime radiance 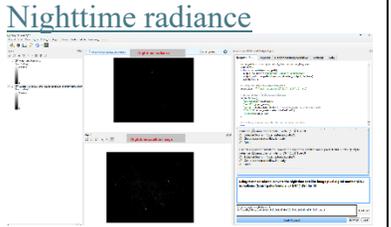 |
| 49 | Generate a raster distance for the fastfood restaurants. | Distance Raster Generation | | 1 | Richland_SC_fastfood.shp | Raster_Distance_Result.png 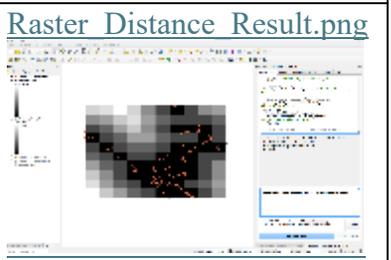 |
| 50 | Generate the roughness index based on the DEM data. | Terrain Roughness Analysis | ['gdal:roughness'] | 1 | Richland_SC_DEM.tif | RoughnessIndex_Result.png 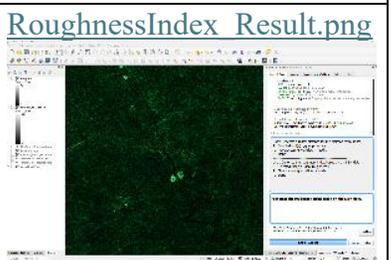 |

| S/N | Task | Category | Tool(s) used | Number of attempts | Dataset used | Screenshot |
|---|---|---|---|---|---|---|
| 51 | Generate the terrain ruggedness index based on the DEM data. | Terrain Analysis. | ['gdal:triterrainruggednessindex'] | 1 | Richland_SC_DEM.tif | terrain_ruggedness_index_Result.png 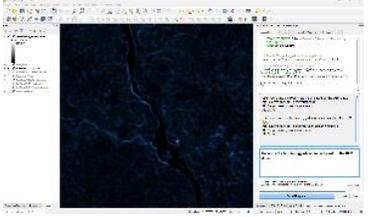 |
| 52 | Reproject the building layer to UTM Zone 17 and then convert the projected data to a raster | Data Management and Rasterization | ['gdal:warpreproject', 'gdal:rasterize'] | 1 | PSU_b.shp | Screenshot 2024-10-24 233849 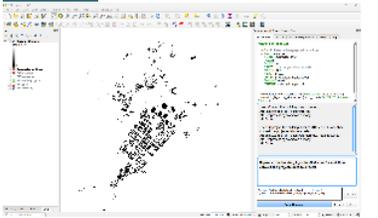 |
| 53 | Use the raster calculator to create a mask showing cells greater than 1000 meters as 1, and cells less than 1000 meters as 0. Use the binary mask expression. | Raster Analysis. | ['qgis:rastercalculator'] | 1 | srtm_41_19.tif | Screenshot.png 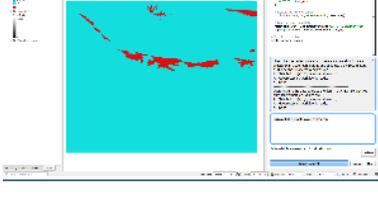 |
| 54 | Set no data value for the pixel values equal to 0 | Raster Data Management. | ['qgis:rastercalculator'] | 1 | elevation_mask.tif | Screenshot 2024-10-24 230407 |

| S/N | Task | Category | Tool(s) used | Number of attempts | Dataset used | Screenshot |
|---|---|---|---|---|---|---|
|  |  |  |  |  |  | 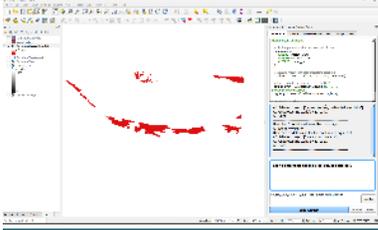 |
| 55 | Convert the mask layer to vector. | Data Conversion | ['gdal:polygonize'] | 1 | elevation_mask_nodata.tif | Screenshot 2024-10-24 231951 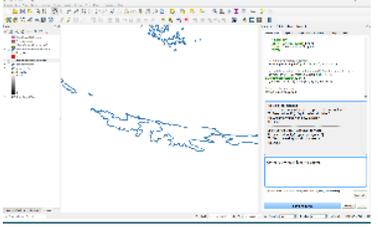 |
| 56 | Save the data as a shapefile | Data Export | ['pdal:convertformat'] | 1 | elevation_mask_nodata.tif | Screenshot 2024-10-24 233219 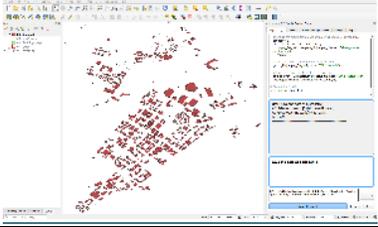 |

| S/N | Task | Category | Tool(s) used | Number of attempts | Dataset used | Screenshot |
|---|---|---|---|---|---|---|
| 57 | Create Thiessen Polygons for the point data | Geostatistical Analysis. | ['native:voronoipolygons'] | 1 | Richland_SC_fastfood.shp | Screenshot 2024-10-25 001058 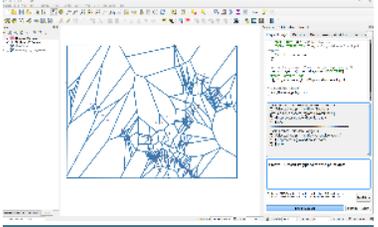 |
| 58 | Please generate a detailed report in HTML to describe the nature of the two datasets based on your observation. In addition, in the report, please suggest some potential applications or analysis scenarios using the two datasets. | Report Generation | ['gdal:gdalinfo', 'native:hillshade'] | 1 | Richland_SC_fastfood.shp  Richland_SC_DEM_UTM17.tif | Screenshot 2024-10-25 001856 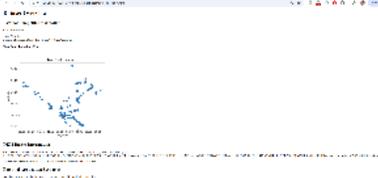 |
| 59 | Generate a HTML table to show the distance matrix in the matrix style. | Matrix Analysis. | ['pandas', 'geopandas'] | 1 | distance_matrix.shp | Screenshot 2024-10-25 002825 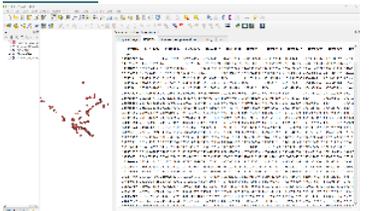 |

| S/N | Task | Category | Tool(s) used | Number of attempts | Dataset used | Screenshot |
|---|---|---|---|---|---|---|
| 60 | Please draw a histogram for the DEM data using plotly | Statistical Analysis | ['Rasterio', 'NumPy', 'Plotly'] | 1 | Richland_SC_DEM_UTM17.tif | Screenshot 2024-10-25 003225 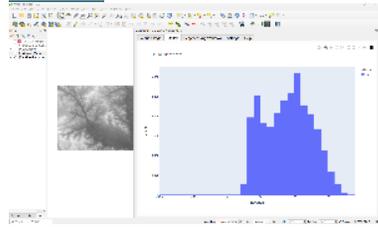 |

## Appendix 4: Intermediate level test cases

| S/N | Task | Category | Tool(s) used | Number of attempts | Dataset used | Screenshot |
|---|---|---|---|---|---|---|
| 1 | Perform the following tasks: 1) Count the fast food restaurants in each county and store the result in a new field named "Count". 2) Calculate the fast food accessibility score for each county as (Count / Population) * 1,000 and store the result in a new field named "Score". 3) | Spatial Analysis | native:countpointsinpolygon, thematic_map_creation | 1 | /Pennsylvania/PA_Fastfoods_XY.shp  /Pennsylvania/PennsylvaniaCounties.shp | Food_Accessibility 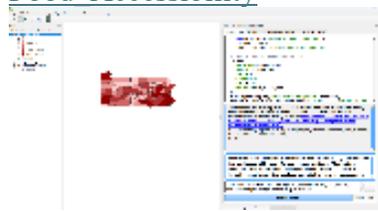 |

| S/N | Task | Category | Tool(s) used | Number of attempts | Dataset used | Screenshot |
|---|---|---|---|---|---|---|
|  | Create a thematic map showing the fast food accessibility score for each county. | | | | | |
| 2 | Perform the following task: 1) Clip the DEM to SC boundary and load the clipped DEM. 2) Show me the histogram for the clipped DEM pixel values. 3) Generate the zonal statistics for every county in SC, focusing on average elevation. 4) Finally create a choropleth map showing the average elevation of each county in SC. | Terrain Analysis | gdal:cliprasterbymasklayer<br><br>seaborn<br><br>native:zonalstatisticsfb | 1 | SouthCarolinaData | SC_DEM Figure 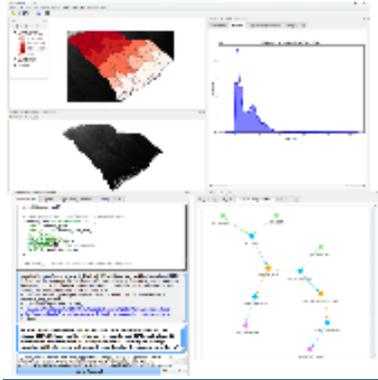 |
| 3 | Can you please generate a HTML report to show the building area for each building? You can look at the columns information, if no building area exists, please calculate yourself. Also make sure to use map projection when calculating area | Attribute Management | native:reprojectlayer<br><br>qgis:exportaddgeometrycolumns | 1 | /Pennsylvania/Penn_State_Buildings.shp | Building report 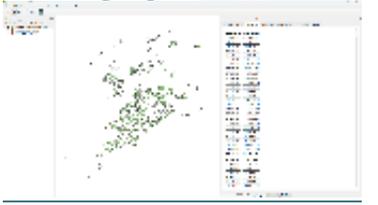 |
| 4 | Perform the following task: 1) Reproject this road vector layer to NAD27/ UTM Zone 10 . 2) Create a new field | Geometric Analysis | gdal:warpreproject<br><br>native:fieldcalculator | 1 | SanFrancisco/Roads | Road_Length |

| S/N | Task | Category | Tool(s) used | Number of attempts | Dataset used | Screenshot |
|---|---|---|---|---|---|---|
|  | that contains the length of the road features. Let the name of the field be "Length". 3) Use the 'Length' column and find how many roads have length shorter 100-meters . |  | native:extractbyattribute |  |  | 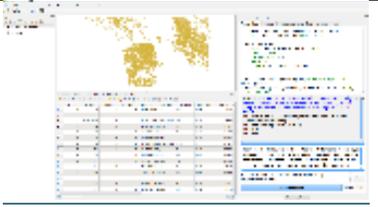 |
| 5 | Calculate the ruggedness index for Pennsylvania, then summarize the ruggedness index values for each counties | Terrain Analysis | ['gdal:triterrainruggednessindex', 'gdal:cliprasterbymasklayer', 'native:zonalstatisticsfb'] | 1 | Pennsylvania/PA_DEM  Pennsylvania/PennsylvaniaCounties | PA_Ruggedness_index 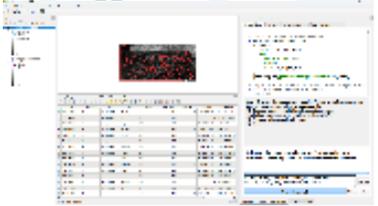 |
| 6 | Add new fields for latitude and longitude to the fast food restaurants layer, then extract the elevation values from the DEM layer and save the results in a new layer. | Data Management | ['native:addxyfields', 'native:rastersampling'] | 1 | Pennsylvania/PA_Fastfoods  Pennsylvania/PA_DEM | Add_XY_to_Fastfood 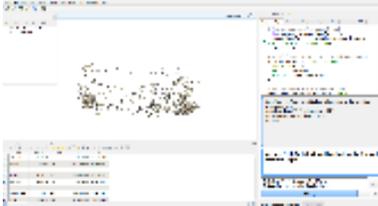 |
| 7 | Merge the four DEMs into a single raster and perform terrain characteristic analysis for Richland County, including slope, aspect, hillshade, terrain ruggedness index (TRI), and topographic Position Index (TPI). | Raster Analysis | pdal:merge, gdal:cliprasterbymasklayer, native:slope, native:aspect native:hillshade, gdal:triterrainruggednessindex, gdal:tpitopographicpositionindex | 1 | Richland | Richland_Raster_Analysis.png |

| S/N | Task | Category | Tool(s) used | Number of attempts | Dataset used | Screenshot |
|---|---|---|---|---|---|---|
|   |   |   |   |   |   | 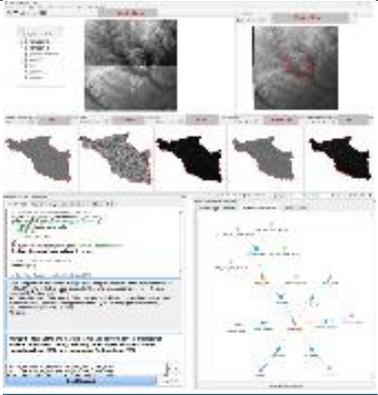 |
| 8 | Count the number of hospitals and number of fastfoods in each county and save the results into a new county layer by adding two fields. | Attribute Query and Data Management | native:joinbylocationsummary | 2 | Pennsylvania/PA_Hospital<br><br>PA_Counties<br><br>PA_Fastfoods | PA_FF_HospitalCount<br>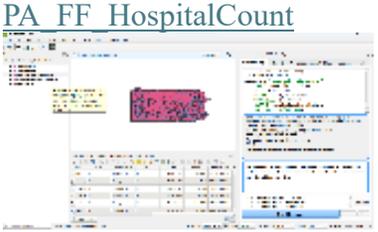 |
| 9 | Reproject this hospital layer to UTM Zone 17, and then create a 1000 meters buffer around each hospital | Geometric Transformation and Buffering Analysis. | 'gdal:warpreproject', 'native:buffer', | 1 | Pennsylvania/PA_Hospital | PA_Hospital_Buffer<br>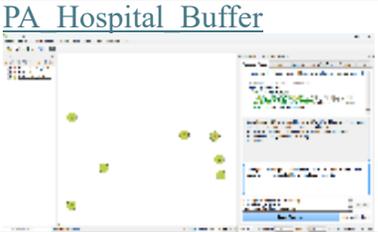 |
| 10 | Project the Nigeria railways into EPSG:26392 map projection, then calculate the total length. | Geometric Transformation and Measurement Analysis. | ['native:reprojectlayer', 'native:sumlinelengths'] | 2 | Nigeria/Nigeria_Major_Roads.shp | 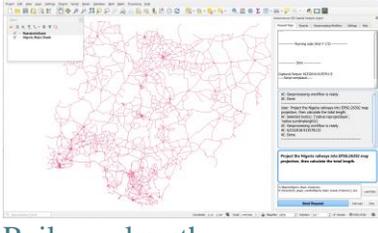<br>Railway_length |

| S/N | Task | Category | Tool(s) used | Number of attempts | Dataset used | Screenshot |
|---|---|---|---|---|---|---|
| 11 | Generate a distance raster map for each grocery store site within 15 km, respectively; the resolution is 50 m; use the "id" column to name the raster files. | Distance Analysis and Raster Generation | | 3 | Walmart_site/Walmart_4_site_EPSG32111.shp | Distance_map.png 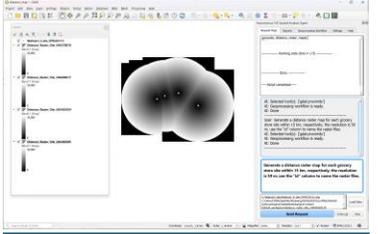 |
| 12 | Select the Census tracts in the Washington area, and then save the tracts that possess the top 10% of the poverty ratio in the selected tracts. | Attribute Query and Statistical Analysis | ['native:extractbyexpression', 'pdal:clip'] | 2 | PovertyData\Washington.shp, PovertyData/Poverty.shp | Washington_poverty.png 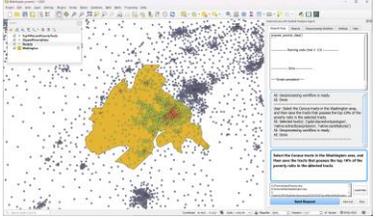 |
| 13 | Create a centroid for each polygon and export the attributes as a table. | Geometric Operations and Data Export | ['native:centroids', 'native:exporttospreadsheet'] | 1 | US_counties | Table_Result 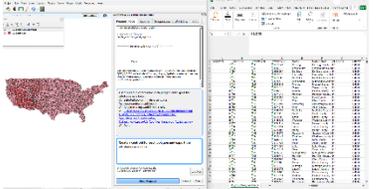 |
| 14 | Conduct a hospital accessibility analysis following these steps: 1) find out the nearest hospital for each Census tract. 2) compute the distance between each tract's centroid and its nearest hospital. 3) create a raster, indicating the distance from | Network Analysis and Distance Analysis | 'native:nearestneighbourhouranalysis', 'native:centroids', 'qgis:distancetonearesthubpoints', 'gdal:rasterize', 'gdal:proximity' | Failed | SanFrancisco_tracts_central, SanFranciscoHospitals | |

| S/N | Task | Category | Tool(s) used | Number of attempts | Dataset used | Screenshot |
|---|---|---|---|---|---|---|
|  | each tract to its nearest hospital. |  |  |  |  |  |
| 15 | Generate a raster map, showing the distances from San Francisco census tracts to the nearest hospitals. The map resolution is 100 m. | Distance Analysis and Raster Generation. | ['gdal:rasterize', 'gdal:proximity'] | 3 | San_Francisco/SanFranciscoHospitals.shp, San_Francisco/SanFrancisco_tracts_central.shp | 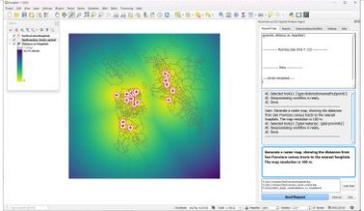 hospital_distance.png |
| 16 | Aggregate the population with obesity for each state. The "obesity" column is the prevalence in percentage. | Statistical Analysis and Data Aggregation. | ['native:aggregate'] | 3 | US_Counties/US_Counties.shp | Obesity_population.png 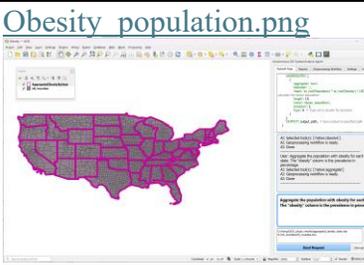 |
| 17 | Find out the top 20 coffee shops in New York City that have the most bus stops or metro entrances within 600 feet. | Proximity Analysis and Spatial Query | ['native:buffer', 'native:countpointsinpolygon'] | 3 | New_York_coffee_shop\bus_metro_stop_EPSG6535.shp, New_York_coffee_shop\New_york_coffee_shops_EPSG6535.shp, | NY_coffee.png 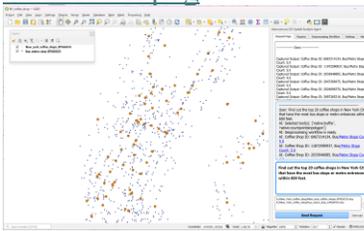 |

| S/N | Task | Category | Tool(s) used | Number of attempts | Dataset used | Screenshot |
|---|---|---|---|---|---|---|
| 18 | Compute the terrain slope for each building: 1) compute the slope; 2) create centroid of building footprints; 3) extract the slope for building centroids. | Terrain Analysis and Geometric Operations | ['native:slope', 'native:centroids', 'native:rastersampling'] | 3 | Rolla_building_footprints, Rolla_DEM_projected | 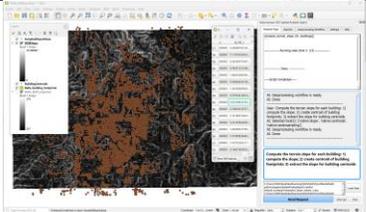 Rolla_building_slope.png |
| 19 | Generate centroids for each building feature, extract the elevation values from the DEM at each building centroid, and export the elevation data to draw a histogram with 50 bins using Seaborn | Geometric Operations and Data Extraction | Native:centroids, native:resampling, seaborn | 1 | Pennsylvania/PSU_DEM  Pennsylvania/Penn_State_Buildings | Penn State Building Elevation 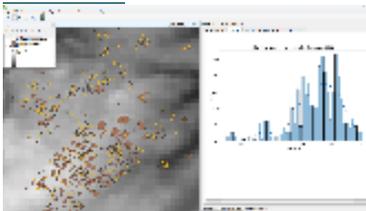 |
| 20 | Show the year of the street view image location. You can consider such steps: 1) extract the year from the "image_date" column; 2) make a thematic map by rendering the extracted year using unique values. | Data Extraction and Thematic Mapping | ['native:calculateexpression', 'thematic_map_creation'] | 3 | Street_view_image\street_image.shp | 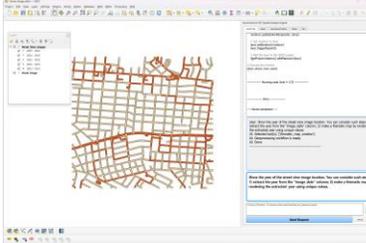 Street_image_date.jpg |
| 21 | Generate an obesity risk behavior index of each county in the USA by analyzing the rate of visits to unhealthy food retailers (such as convenience store, alcoholic drinking places, and limited service | Index Creation and Thematic Mapping | native:joinattributestable, qgis:advancedpythonfieldcalculator, thematic_map_cretion | 2 | Obesity_Risk_Score/US_Counties  Obesity_Risk_Score/Place_visit_rates | Obesity_Risk_Figure_New_New 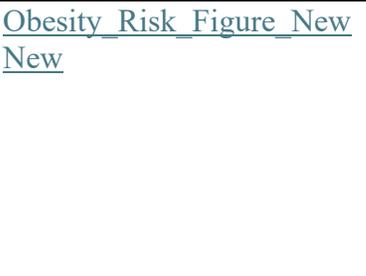 |

| S/N | Task | Category | Tool(s) used | Number of attempts | Dataset used | Screenshot |
|---|---|---|---|---|---|---|
|  | restaurant) and the visit rate to places that support physical activity (e.g., sports centers, parks, fitness centers). Visualize the results in a thematic map to highlight the obesity risk behavior index across counties. |  |  |  |  | 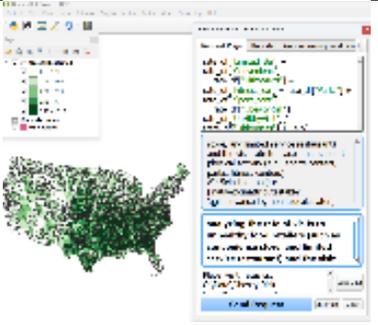 |
| 22 | You need to compute the location error between the estimated tree trunk and the ground truth: 1) for each ground truth location, find out its nearest estimated location within 3 meters; the nearest distance is the error. 2) compute the average error for ground truth trunks that have the associated estimated location. | Accuracy Assessment and Distance Analysis | ['qgis:distancetonearesthubpoints'] | Failed. | Street_view_image/trunk_location_estimated_EPSG6569.shp, Street_view_image/trunk_location_ground_truth_EPSG6569.shp |  |
| 23 | Link the street view image location points into a path. The image was obtained using a mapping vehicle. All points represent the location of a street view image. Each image has an identifier in "panoId" column; the column "link_f_pan" indicates the panoId of the | Network Analysis and Data Integration | ['GeoPandas', 'Shapely'] | 3 | Street_view_image\street_image.shp | 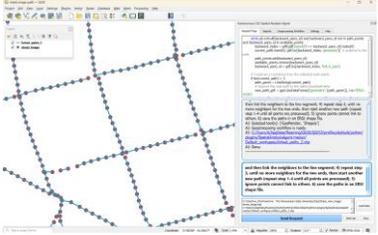
path_linking |

| S/N | Task | Category | Tool(s) used | Number of attempts | Dataset used | Screenshot |
|---|---|---|---|---|---|---|
|  | neighbor image in the forward direction, the column "link_b_pan" indicates the panoId of the neighbor image in the backward direction. You can link the location points by: 1) randomly pick up a point; 2) extract its forward and backward neighbors, link them together to form a line segment; 3) find the neighbors of the two ends of the segment, and then link the neighbors to the line segment; 4) repeat step 3, until no more neighbors for the two ends, then start another new path (repeat step 1-4 until all points are processed); 5) ignore points cannot link to others. 6) save the paths in an ERSI shape file. |  |  |  |  |  |
| 24 | Join the U.S. Census block group population table to the block group boundaries. Note that you must first create the full block group FIPS for the table by | Data Join and Attribute Management | ['pandas', 'native:fieldcalculator', 'native:joinattributestable'] | 1 | Rolla_watershed/Rolla_2021_population_block_group.csv, Rolla_watershed/ Rolla_downtown_bl | Attribute_join.png |

| S/N | Task | Category | Tool(s) used | Number of attempts | Dataset used | Screenshot |
|---|---|---|---|---|---|---|
|  | merging state, county, tract, and block group short FIPSs. |  |  |  | ock_groups_2021.shp | 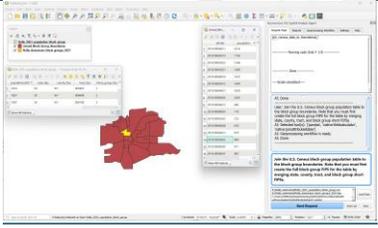 |
| 25 | Allocate the Census block group population to the building footprint according to the building area. For example, in a block group of 30 residents, there are two buildings with areas of 100 square meters and 200 square meters, respectively; you need to allot 10 residents and 20 residents to the two buildings accordingly. | Data Interpolation | ['native:joinattributestable', 'qgis:exportaddgeometrycolumns', 'native:fieldcalculator'] | Failed. | Rolla_watershed/Rolla_building_footprints.shp, Rolla_watershed/Rolla_downtown_block_groups_2019.shp |  |
| 26 | Use the county boundary to clip the land cover raster for each county in Wyoming State, US. These are steps for your reference: 1) Select the county boundary of a county. 2) Clip the land cover raster using the county boundary. 3) Save each county raster as TIF. | Raster Clipping and Data Export | ['native:extractbyattribute', 'gdal:cliprasterbymasklayer', 'gdal:translate'] | 3 | Landcover\US_Counties_projected.shp, Landcover\nlcd_2021_land_cover_l48_20230630.img | 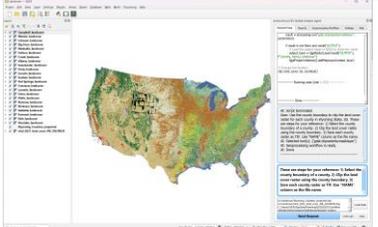 county_clip_landcover.png |

| S/N | Task | Category | Tool(s) used | Number of attempts | Dataset used | Screenshot |
|---|---|---|---|---|---|---|
|  | Use "NAME" column as the file name. |  |  |  |  |  |
| 27 | Clip the image using the extent box of the shape file using these steps: 1) generate the extent box of the shape file; 2) clip and save the image. | Raster Clipping and Geometric Operations | ['native:extenttolayer', 'gdal:cliprasterbyextent'] | Failed. | NoiseMap\Columbia_image_points_ESRI102039.shp, NoiseMap/SC_road_noise_2018.tif |  |
| 28 | North Carolina will be receiving funding from a car company for building electric charging stations. A group of experts have been asked to determine the factors and weights for locating the new EV charging stations in NC. These experts have determined the most important factors in ranking candidate locations for these new facilities (Data.csv). These experts have also assigned weights to each of the factors. Complete the nominal group process technique by calculating the final weights in the csv file | Weighted Analysis and Decision Support | ['pandas'] | 1 | Data.csv | An HTML output |

| S/N | Task | Category | Tool(s) used | Number of attempts | Dataset used | Screenshot |
|---|---|---|---|---|---|---|
|  | on a 0 to 1.0 point scale using the experts factors and their individual weights. |  |  |  |  |  |
| 29 | Using the input DEM layer, generate a watershed with the outlet point specified by the 'Outlet' shapefile. Follow these steps to complete the analysis:<br><br>1. Apply a fill operation on the DEM to remove sinks and ensure proper flow.<br>2. Perform a flow direction analysis based on the filled DEM data.<br>3. Calculate the watershed using the flow direction output and the specified outlet point.<br>4. Run a flow accumulation analysis based on the flow direction output. | Hydrological Analysis and Watershed Modeling | ['native:fillnodata', 'native:aspect', 'grass7:r.watershed', 'qgis:rastercalculator', 'grass7:r.stream.extract'] | Failed | n40_w078_1arc_v3.tif;<br><br>Outlet.shp |  |

| S/N | Task | Category | Tool(s) used | Number of attempts | Dataset used | Screenshot |
|---|---|---|---|---|---|---|
| | 5. Finally, generate streamlines from the flow accumulation output, considering only areas with flow accumulation values greater than 200. | | | | | |
| 30 | Use Inverse Distance Weighted (IDW) interpolation to create a DEM using the given point data. The elevation column is "elevatio_1". The raster resolution is 10 m. Please only use the points having the "image_date" column values in year 2013 and after. | Spatial Interpolation and Raster Generation | ['idw_interpolation'] | 1 | Street_tree/Richland_downtown_street_view_image_location_EPSG6569.shp | 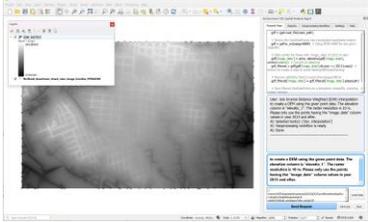 IDW.png |

# Appendix 5: Advanced level test cases

| S/N | Task | Category | Tool(s) used | Number of attempts | Dataset used | Screenshot |
|---|---|---|---|---|---|---|
| | | | | | | |

| 1 | Identify the street tree count within 20-meter buffer for each street. | Proximity Analysis and Spatial Query | ['gdal:buffervectors', 'native:joinbylocationsummary'] | 2 | Street_tree/Street_Tree_EPSG6852  Street_tree/Roads_Potland_EPSG6852_subset | Tree_count.png |
|---|---|---|---|---|---|---|
| 2 | Compute the total highly developed area of Orange County, NY, according to the land cover raster. The pixel value of highly developed area is 24. The unit of area is square kilometer. | Area Calculation and Raster Analysis | ['gdal:cliprasterbymasklayer', 'qgis:rastercalculator', 'native:rasterlayerstatistics'] | 3 | nlcd_2021_land_cover_I48_20230630.tif  NewYork_Counties_Projected | Developed_area.png |
| 3 | Compute the road lengths inside each protected region. The identifier of projected region is "fid". | Spatial Analysis and Length Calculation | ['native:intersection', 'qgis:exportaddgeometrycolumns', 'native:aggregate', 'pandas', 'tabulate'], ['native:joinbylocationsummary', 'native:exporttospreadsheet'] | Failed. | Multicriteria Overlay Analysis/roads_polyline.shp,  Multicriteria Overlay Analysis/Protected_region.shp | |
| 4 | From PA_Counties dataset select only those counties that the number of data from PA_DATA_EPSG4326 are completely within their boundaries but the count of point is less than 10. Then | Spatial Query and Data Extraction | ['native:countpointsinpolygon', 'native:extractbyattribute'] | 1 | PA counties shapefile ;  PA_Data_EPSG4326 | Level3_4 |
| | extract these counties and load them into QGIS. | | | | | |
| 5 | I want you to measure the length of boundary of each county in km based on this input data. Then display your result in an HTML file. | Boundary Measurement and Data Export | ['qgis:exportaddgeometrycolumns', 'Pandas library'] | 1 | PA counties | PA_Counties_BoundaryLengths.png 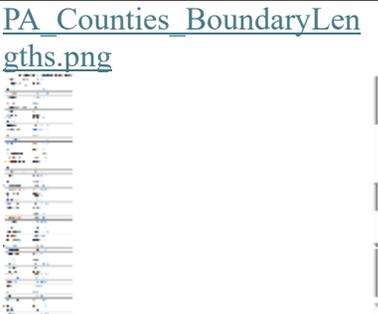 |
| 6 | I want you to find all roads that crossed the center county and then extract them from tl_2019_42_prisecroads. | Spatial Query and Data Extraction | ['qgis:selectbyattribute', 'native:extractbyattribute', 'native:extractbylocation'] | 2 | TL_2019_42_PRISECROADS | Level3_6 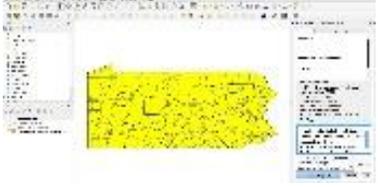 |
| 7 | For each county, 1) find the weather stations inside the county, and 2) list the weather station names inside the county. | Spatial Query and Data Aggregation | ['native:joinattributesbylocation', 'native:extractbyattribute'] | 2 | SC_weatherstations SC_county_boundaries | 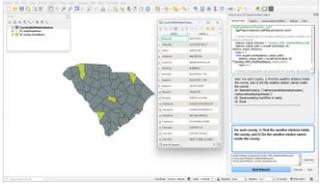 weather_station.png |
| 8 | I would like to identify which counties in Pennsylvania are suitable for planting more trees, using annual rainfall as a key parameter. Counties receiving more than 2.5 inches of rain per year should be considered suitable for tree planting. | Suitability Analysis and Statistical Assessment | ['native:extractbyattribute', 'qgis:basicstatisticsforfields'] | 1 | Level3_30.shp | Tree_Plalnting_Suitability.png 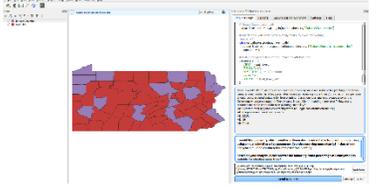 |
| | Based on your analysis, please answer the following: what percentage of Pennsylvania is suitable for planting more trees? How many counties meet the criteria for tree planting suitability? | | | | | |
| 9 | Calculate the population that live in the block groups having hazard wastes. | Population Analysis and Spatial Query | ['native:reprojectlayer', 'native:joinattributesbylocation'] | Failed. | HW_Sites_EPSG4326/North_Carolina_blockgroup_boundaries.shp, HW_Sites_EPSG4326/HW_Sites_EPSG4326.shp, HW_Sites_EPSG4326/blockgroup_population_north_carolina.csv | 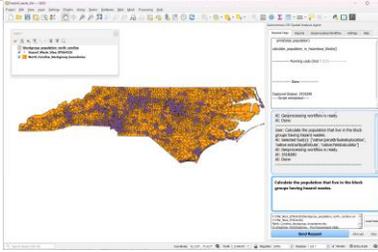Hazard_waste.png |
| 10 | Based on this input data, identify areas suitable for solar farm installation based on the slope and proximity to power line (RandomPoint). The slope should be less than 5 degrees, and the distance to the nearest power line (RandomPoint) should be within 500 meters. | Suitability Analysis and Proximity Analysis | ['native:slope', 'qgis:rastercalculator', 'native:buffer', 'native:polygonize', 'pdal:clip'] | 2 | n40_w078_1arc_v3.tif | Level3_9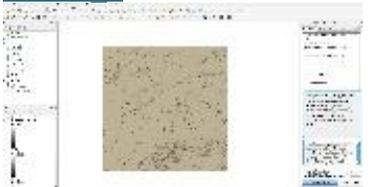 |

| 11 | For each school in Columbia, calculate the length of sidewalks within 500 meters. | Proximity Analysis | ['gdal:buffervectors', 'gdal:clipvectorbypolygon', 'native:sumlinelengths'] | 1 | Columbia_school_walkability/Columbia_sidewalk_EPSG6569.shp, Columbia_school_walkability/Columbia_schools_EPSG6569.shp | school_sidewalk_length |
| --- | --- | --- | --- | --- | --- | --- |
| 12 | I want you to join these two files and create one file that counties contain HIV rates based on this dataset 'HIV Point'. Then generate a symbology map based on HIV rate. | Data Join and Thematic Mapping | ['native:joinattributesbylocation', 'thematic_map_creation'] | 1 | HIV shapefile; US counties | Level3_14 |
| 13 | Could you analyze and visualize the fast food accessibility score for each county based on the number of fast food restaurants and population using a thematic map with blue graduated colors. Then, analyze the correlation between the county-level obesity rate and the fast food accessibility score by drawing a scatter plot with a regression line | Statistical Analysis and Thematic Mapping | ['thematic_map_creation, 'seaborn'] | 1 | PA_Fastfoods_XY, Pennsylvaniacounties | Fastfood_accessibility score |

| 14 | Based on this provided DEM data and another observation point (Random_Point), I want you to conduct a viewshade analysis from the observation point. Set a maximum radius for the analysis of 5 kilometers. | | Selected tool(s): ['gdal:viewshed'] | 1 | n40_w078_1arc_v3; Random_Point | Level3_26 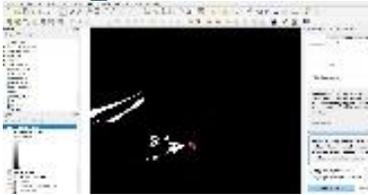 |
| --- | --- | --- | --- | --- | --- | --- |
| 15 | Please conduct a visibility analysis using the selected datasets. Set the following parameters for the analysis: Observation Offset (Cell Towers): 30 meters. Surface Offset: 1.60 meters. Outer Radius: 2000 meters. After completing the analysis, answer the following questions: How many buildings are visible from a single cellular tower? What percentage of all buildings are visible from five cellular towers? | Viewshed Analysis and Terrain Analysis | ['gdal:viewshed', 'Intersect'] | Failed | Viewshade_analysis file | |
| 16 | Extract the DEM values at the street view images' locations, and then compare them with the elevation in the image's metadata ("elevatio_1" column), print out the descriptive statistics of the difference between the | Data Extraction and Statistical Analysis | ['native:rastersampling', 'gdal:gdal2xyz', 'grass7:r.univar', 'native:importintopostgis'] | 1 | Street_view_image/street_image.shp, Street_view_image/Richland_DEM_EPSG_6569.tif | street_image_DEM.png, 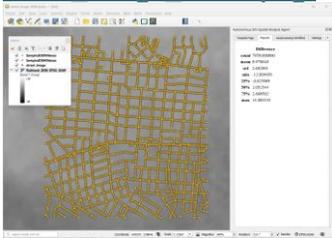 |
| | DEM and metadata in an HTML file. | | | | | |
| 17 | Propose a new Walmart store location, which has similar distances to the existing four locations, while near the roads within 300 meters. | Site Suitability Analysis and Proximity Analysis | ['gdal:buffervectors', 'qgis:distancematrix', 'native:extractwithindistance', 'native:intersection'] | Failed | Walmart_site/Walmart_4_site_EPSG32111.shp, Walmart_site/Motorway_EPSG32111.shp | |
| 18 | Generate the Normalized Difference Vegetation Index (NDVI) of Akure from these satellite imageries. | Raster Analysis | ['qgis:rastercalculator', 'native:reprojectlayer', 'gdal:cliprasterbymasklayer'] | 1 | Landsat 8 imageries, Akure_boundary | NDVI.png 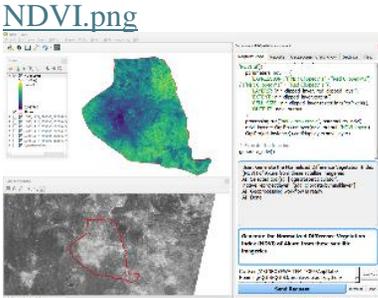 |
| 19 | Use the given checkpoint locations, extract the values of the two DEM files, and then compute the RMSE. | Data Extraction | ['native:rastersampling'] | 1 | Street_tree/idw_output.tif, Street_tree/Checkpoints_Columbia.shp, Street_view_image/Richland_DEM_EPSG_6569.tif | DEM_RMSE.png 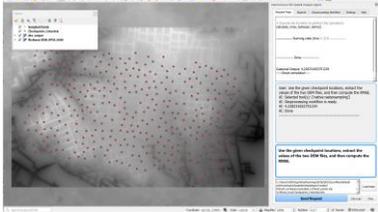 |
| 20 | Can you help me identify the areas that are suitable for camping. I want the aspect of the area to be greater than 100, and the distance to a closest road not exceeding 1000 meters. | Suitability Analysis and Proximity Analysis | ['native:aspect', 'qgis:rastercalculator', 'gdal:rasterize', 'gdal:proximity', 'qgis:rastercalculator'] | Failed | PA_DEM.tif, PA_Roads | |

| | | | | | |
|---|---|---|---|---|---|
| | | | | | |